\documentclass[graybox]{svmult}

\usepackage{type1cm}        
\usepackage{makeidx}         
\usepackage{graphicx}        
\usepackage{comment} 
\usepackage{multicol}        
\usepackage[bottom]{footmisc}

\usepackage{booktabs}
\usepackage{longtable}

\usepackage{newtxtext}       %
\usepackage[varvw]{newtxmath}       
\usepackage{algorithm}
\usepackage{algorithmic}

\makeindex             

\begin{document}

\title*{ Multi-Task Learning Based on Support Vector Machines and Twin Support Vector Machines: A Comprehensive Survey}
\titlerunning{Multi-Task Learning Based on Support Vector Machines} 

\author{ 	 Fatemeh Bazikar \orcidID{0000-0002-7497-935X}  and \\Hossein Moosaei \orcidID{0000-0002-0640-2161} and \\Atefeh Hemmati \orcidID{0000-0002-6017-6820}   and \\
 Panos M. Pardalos \orcidID{0000-0001-9623-8053}}

\authorrunning{F. Bazikar et al.}

\institute{
	Fatemeh Bazikar \at Department of Computer Science, Faculty of Mathematical Sciences, Alzahra University, Tehran, Iran \\ \email{F.Bazikar@alzahra.ac.ir}
	\and Hossein Moosaei \at Department of Informatics, Faculty of Science, Jan Evangelista Purkyne University, Ústí nad Labem, Czech Republic \\ \email{hossein.moosaei@ujep.cz}
	\and Atefeh Hemmati \at Department of Computer Engineering, Science and Research Branch, IAU, Tehran, Iran \\ \email{atefeh.hemmati@srbiau.ac.ir}
	\and Panos Pardalos \at Department of Industrial and Systems Engineering, University of Florida, USA \\ \email{pardalos@ise.ufl.edu}
}
%
%
\maketitle

\abstract{ Multi-task learning (MTL) provides an effective framework for jointly training models across multiple related tasks, enabling them to leverage shared structure and enhance overall generalization. Although deep learning is widely used in MTL, Support Vector Machines (SVM) and Twin Support Vector Machines (TWSVM) continue to provide reliable and interpretable solutions in data-scarce or high-dimensional scenarios. This chapter presents an in-depth survey of MTL approaches based on SVM and TWSVM formulations. We examine key methodological developments that incorporate shared representations, regularization across tasks, and structural coupling strategies. A special focus is placed on emerging extensions of TWSVM to multi-task settings, an area that has received relatively limited attention but holds substantial promise. We further compare these models in terms of theoretical properties, optimization strategies, and empirical performance across application domains. The chapter concludes by identifying major research gaps and outlining future directions toward building interpretable, scalable, and reliable MTL frameworks based on margin-based learning theory.
}

Multi-task learning (MTL) has emerged as a powerful paradigm in machine learning, enabling the simultaneous modeling of multiple related tasks to leverage shared information, thereby enhancing generalization, efficiency, and robustness compared to traditional single-task learning approaches. By exploiting inter-task relationships, MTL facilitates knowledge transfer, reducing overfitting in data-scarce scenarios and improving predictive performance across diverse applications, including natural language processing, computer vision, and bioinformatics. While deep learning has dominated recent MTL advancements due to its flexibility in modeling complex data, margin-based methods such as Support Vector Machines (SVMs) and their variant \cite{ref1,ref2,ref3,ref4}, Twin Support Vector Machines (TWSVMs)\cite{ref11}, remain highly relevant for their theoretical rigor, interpretability, and effectiveness in high-dimensional or small-sample settings. This chapter provides a comprehensive survey of MTL approaches grounded in SVM and TWSVM frameworks, with a focus on their methodological developments, optimization strategies, and practical applications.

The motivation for integrating SVM and TWSVM with MTL lies in their complementary strengths. SVMs, rooted in statistical learning theory, maximize the margin between classes to ensure robust generalization, making them particularly suitable for scenarios where data is limited or noisy \cite{ref5}. Their convex optimization formulations and kernel-based flexibility enable effective handling of both linear and non-linear relationships. TWSVMs, an evolution of SVMs introduced by \cite{ref11}, enhance computational efficiency by solving two smaller quadratic programming problems (QPPs) for non-parallel hyperplanes, each tailored to one class. This approach not only accelerates training but also adapts well to imbalanced or complex data sets, making TWSVM a promising candidate for MTL extensions.

The need for MTL arises in domains where tasks are inherently related, such as predicting protein functions across species, classifying images from multiple categories, or forecasting outcomes for related financial instruments. Traditional single-task learning often fails to capture these relationships, leading to suboptimal models, especially when data per task is limited. MTL addresses this by sharing knowledge across tasks, either through feature representations, model parameters, or structural constraints. However, the application of SVM and TWSVM in MTL has been underexplored compared to deep learning approaches, despite their advantages in interpretability, computational efficiency for smaller data sets, and theoretical guarantees. This chapter aims to fill this gap by systematically reviewing SVM- and TWSVM-based MTL methodologies, highlighting their contributions and identifying areas for future research.

The scope of this chapter encompasses the following objectives:
\begin{itemize}
	\item To provide a detailed examination of the theoretical foundations of SVM and TWSVM, establishing the groundwork for their MTL extensions.
	\item To categorize and analyze SVM-based MTL approaches, including parameter-sharing, kernel-based, and constraint-based methods, with emphasis on their optimization techniques and performance.
	\item To explore recent advancements in TWSVM for MTL, an area with limited but growing literature, focusing on its potential to address scalability and task heterogeneity.
	\item To compare these approaches in terms of theoretical properties, computational efficiency, and empirical performance across diverse application domains.
	\item To identify research gaps and propose directions for developing scalable, interpretable, and robust MTL frameworks using margin-based learning principles.
\end{itemize}
This chapter targets researchers and practitioners in machine learning, offering a comprehensive resource for understanding and advancing SVM- and TWSVM-based MTL.
\section{Fundamentals of Multi-Task Learning}
MTL represents a paradigm in machine learning that seeks to improve the performance of multiple related tasks by leveraging shared information across them. Unlike traditional single-task learning, where each task is modeled independently, MTL exploits the inherent relationships between tasks to enhance generalization, efficiency, and robustness. This section provides a foundational overview of MTL, beginning with its definitions and learning settings, followed by a categorization of approaches, and concluding with a discussion of its advantages and challenges.

\subsection{Definitions and Learning Settings}
At its core, MTL can be formally defined as the process of learning a set of $T$ related tasks simultaneously, where each task $t$ (for $t = 1, \dots, T$) is associated with its own data set $\mathcal{D}_t = \{ (\mathbf{x}_{t,i}, y_{t,i}) \}_{i=1}^{n_t}$. Here, $\mathbf{x}_{t,i} \in \mathbb{R}^d$ denotes the input features for the $i$-th sample in task $t$, $y_{t,i}$ is the corresponding label (which could be continuous for regression or discrete for classification), and $n_t$ is the number of samples for that task. The goal is to learn a set of task-specific models $f_t: \mathbb{R}^d \to \mathcal{Y}_t$ (where $\mathcal{Y}_t$ is the output space for task $t$) that benefit from shared knowledge, often parameterized by a common structure or constraints.

The learning settings in MTL vary depending on the nature of the individual tasks and the availability of data. Drawing from established frameworks \cite{ref6,ref7,ref8,ref52,ref53,ref54}, key settings include:
\begin{itemize}
	\item\textbf{Multi-Task Supervised Learning:} All tasks are supervised, with labeled data available for each task. This is the predominant setting, encompassing classification, regression, or related predictive tasks. For example, in natural language processing, tasks might involve part-of-speech tagging and named entity recognition applied to the same corpus, where joint learning leverages shared information to enhance accuracy .
	\item\textbf{ Multi-Task Semi-Supervised Learning:} Tasks incorporate both labeled and unlabeled data. Here, shared representations derived from labeled tasks can regularize or guide learning in tasks with limited labels, mitigating data scarcity issues common in domains like medical image analysis.
	\item\textbf{ Multi-Task Unsupervised Learning:} Tasks focus on discovering patterns without labels, such as clustering or density estimation. MTL in this context promotes consistency across tasks, for instance, by learning shared latent spaces that capture underlying structures in multiple data sets.
	\item\textbf{ Multi-Task Active Learning:} This setting extends semi-supervised approaches by actively selecting unlabeled instances for labeling. Shared knowledge across tasks informs query strategies, optimizing label acquisition to improve overall performance.
	\item \textbf{ Multi-Task Reinforcement Learning:} Tasks involve sequential decision-making to maximize cumulative rewards. MTL enables agents to transfer policies or value functions between related environments, accelerating learning in scenarios like robotics or game AI.
	\item \textbf{ Multi-Task Online Learning:} Tasks process data sequentially, adapting models in real-time. This is useful for streaming applications, where MTL exploits inter-task relations to handle non-stationary data distributions.
	\item \textbf{ Multi-Task Multi-View Learning}: Tasks operate on multi-view data, where each instance is described by multiple feature sets (e.g., text and images). MTL enforces complementarity across views, enhancing robustness in multi-modal contexts.
\end{itemize}
MTL also relates to transfer learning and domain adaptation as special cases. Transfer learning typically uses one or more source tasks to aid a target task, whereas MTL symmetrically learns all tasks jointly. Domain adaptation addresses distribution shifts across tasks, often as a subset of MTL. Settings can be homogeneous (tasks share input/output spaces) or heterogeneous (tasks differ in modalities or label types, e.g., image classification and text sentiment analysis). A core assumption is task relatedness; forcing unrelated tasks into shared models can lead to negative transfer, degrading performance.
\subsection{Categories of Multi-Task Learning Approaches}
MTL approaches are categorized based on mechanisms for inducing knowledge sharing among tasks, primarily in supervised settings \cite{ref7,ref8}. These include feature-based, parameter-based, and instance-based methods, providing a structured lens for the field's methodologies.
\begin{itemize}
	\item \textbf{ Feature-Based Approaches:} These methods emphasize learning shared feature representations across tasks, derived from original features to capture task-invariant information while permitting task-specific adaptations. Sub-approaches include feature transformation, feature selection, and deep learning. For instance, the multi-task feature learning (MTFL) method employs $l_{2,1}$ regularization to enforce row-sparsity in the parameter matrix, promoting a common feature subset. This category excels in high-dimensional data, reducing overfitting by aggregating task-specific information.
	\item \textbf{ Parameter-Based Approaches:} Sharing occurs through model parameters, which are related via priors, regularizations, or decompositions. Sub-approaches encompass low-rank methods, task-clustering, task-relation learning, dirty decomposition, and multi-level decomposition. A representative example is the multi-task relationship learning (MTRL) method, which learns a positive semidefinite task covariance matrix through matrix-variate normal priors. Bayesian extensions, such as hierarchical models, enable probabilistic inference of task relations.
	\item \textbf{ Instance-Based Approaches:} These leverage instances from all tasks to build learners for each, often via weighting or co-training. For example, methods may pool instances across related tasks or use co-regularization to align predictions on shared unlabeled data.
\end{itemize}
These categories are not exhaustive or mutually exclusive; hybrid models often integrate elements, such as combining feature sharing with parameter constraints, to optimize performance in diverse applications.
\subsection{ Advantages and Challenges}
Multi-Task Learning is a machine learning paradigm that aims to solve multiple learning tasks simultaneously by exploiting useful information shared among them. As originally emphasized by Caruana, “MTL is an approach to inductive transfer that improves generalization by using the domain information contained in the training signals of related tasks as an inductive bias. It does this by learning tasks in parallel while using a shared low-dimensional representation; what is learned for each task can help other tasks be learned better” \cite{ref47}. The fundamental assumption underlying MTL is that tasks are not entirely independent; rather, they are often correlated or share common structures. By leveraging these correlations, MTL can significantly enhance model performance compared to training each task independently.

The usefulness of MTL becomes particularly evident in high-dimensional problems with limited training data. In such scenarios, traditional single-task learning methods struggle to reliably estimate model parameters, leading to poor generalization. This issue is critical in domains such as medical image analysis or bioinformatics, where labeled data are costly and labor-intensive to obtain \cite{ref50,ref51}. By transferring knowledge across related tasks, MTL mitigates data scarcity and provides more robust solutions.

MTL provides several key advantages over independent single-task learning \cite{ref7,ref8}:
\begin{itemize}
	\item
	Enhanced Generalization: By exploiting inter-task information, MTL acts as an implicit regularizer, improving prediction accuracy, especially in label-scarce domains such as health informatics.
	\item
	Computational Efficiency: Joint optimization enables shared computations (e.g., feature extraction), making MTL suitable for large-scale applications like recommender systems and web search.
	\item
	Robustness to Noise and Sparsity: Data-rich tasks can support noisier or incomplete ones, thereby increasing overall reliability.
	\item
	Broad Applicability: MTL has demonstrated empirical and theoretical success in diverse fields such as computer vision, speech processing, and ubiquitous computing, enabling knowledge transfer mechanisms that resemble human learning.
	\item
	Potential Benefits Beyond Related Tasks: Although most effective when tasks share significant commonalities, studies have shown that MTL can also provide advantages even when tasks are less closely related \cite{ref49}.
\end{itemize}
Despite these benefits, MTL introduces several challenges:
\begin{itemize}
	\item
	Negative Transfer: Unrelated or weakly correlated tasks may introduce interference and degrade performance, necessitating adaptive mechanisms such as task clustering or outlier detection.
	\item
	Scalability Issues: As the number of tasks or the volume of distributed data grows, optimization complexity increases, requiring efficient parallel and distributed frameworks.
	\item
	Data Imbalance: Differences in task-specific data set sizes can bias models toward dominant tasks, which motivates solutions like dynamic loss weighting or data resampling strategies.
	\item
	Evaluation Complexity: Assessing multi-task performance requires multi-objective evaluation frameworks, often based on aggregated metrics or Pareto optimality to capture trade-offs across tasks.
\end{itemize}
In summary, MTL provides a principled approach to addressing challenges in high-dimensional, data-scarce environments by leveraging inter-task dependencies. While its advantages in generalization, efficiency, and robustness are well-established, careful consideration of its limitations—such as negative transfer and scalability—remains crucial. These foundational insights underpin more advanced formulations, including those based on SVMs, which will be discussed in subsequent sections.
\section{Support Vector Machines: A Primer}
Support vector machines represent a cornerstone family of supervised machine learning algorithms, grounded in statistical learning theory and convex optimization principles. Originally formulated for binary classification by Cortes and Vapnik \cite{ref5}, SVMs have been extended to regression, multi-class problems, and beyond, offering robust solutions across diverse domains such as bioinformatics, text categorization, and computer vision \cite{ref9}. The core objective of SVMs is to identify an optimal hyperplane that maximizes the margin—the perpendicular distance to the nearest training points (support vectors)—between classes, thereby enhancing generalization and mitigating overfitting. This primer elucidates the foundational formulation of SVMs in linear and nonlinear contexts, encompassing primal and dual representations, kernel-induced mappings, and regularization strategies. Although deep learning has overshadowed SVMs in ultra-high-dimensional regimes, their interpretability, theoretical bounds on generalization error, and efficacy on modestly sized data sets ensure their enduring relevance \cite{ref9,ref10}.
\subsection{ Linear SVM Formulation}
The linear SVM paradigm assumes data separability in the input space, targeting binary classification. Given a training data set $\mathcal{D} = \{( \mathbf{x}_i, y_i )\}_{i=1}^n$, where $\mathbf{x}_i \in \mathbb{R}^d$ denotes feature vectors and $y_i \in \{-1, +1\}$ the associated labels, the task is to derive a hyperplane $\mathbf{w} \cdot \mathbf{x} + b = 0$, with $\mathbf{w} \in \mathbb{R}^d$ as the normal vector and $b \in \mathbb{R}$ as the bias, that bifurcates the classes while maximizing the margin.

Hard-Margin Primal Formulation: For linearly separable data, the optimization seeks to minimize the weight vector's norm, which inversely scales with the margin (specifically, the full margin spans $2 / \|\mathbf{w}\|$):
\begin{align}
	\min_{\mathbf{w}, b} \quad & \frac{1}{2} \|\mathbf{w}\|^2  \\
	\text{s.t.} \quad & y_i (\mathbf{w} \cdot \mathbf{x}_i + b) \geq 1, \quad \forall i = 1, \dots, n. \nonumber
\end{align}

\begin{figure}
	\centering
	\includegraphics[scale=0.5]{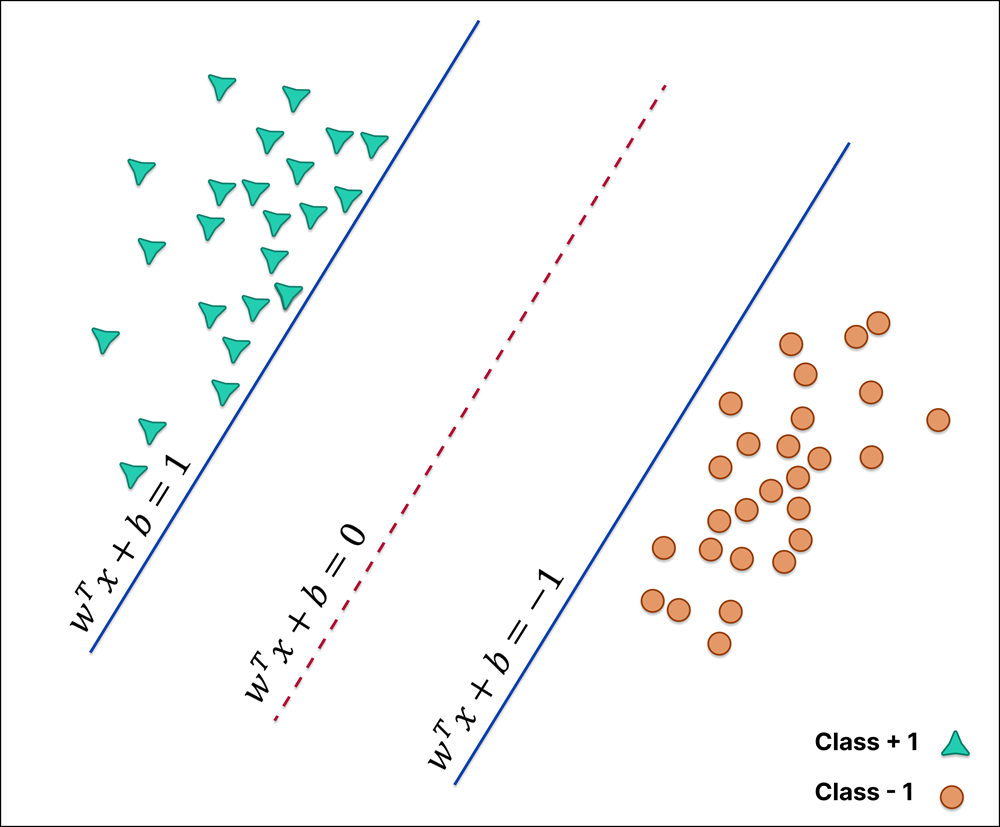}
	\caption{Geometric illustration of the SVM framework.}
	\label{fig:SVM}
\end{figure}
Figure \ref{fig:SVM} illustrates the classification of two classes using SVM.
These inequalities mandate that each point resides on the appropriate side of the hyperplane, at a normalized distance of at least unity from the decision boundary. The support vectors—those instances saturating the constraints—uniquely determine the hyperplane, underscoring SVMs' sparsity and efficiency \cite{ref9,ref10}.
In practice, data sets often exhibit noise or overlap, rendering hard-margin infeasibility. The soft-margin variant introduces slack variables $\xi_i \geq 0$ to accommodate misclassifications or margin violations, balancing margin maximization against empirical error via a regularization parameter $C > 0$:
\begin{align}
	\min_{\mathbf{w}, b, \xi} \quad & \frac{1}{2} \|\mathbf{w}\|^2 + C \sum_{i=1}^n \xi_i \\
	\text{s.t.} \quad & y_i (\mathbf{w} \cdot \mathbf{x}_i + b) \geq 1 - \xi_i, \quad \forall i = 1, \dots, n.\nonumber
\end{align}
This quadratic program, solvable via convex optimization (e.g., sequential minimal optimization (SMO)\cite{ref46}), yields a decision function $f(\mathbf{x}) = \operatorname{sign}(\mathbf{w} \cdot \mathbf{x} + b)$ for prediction, with $C$ trading off between structural risk and empirical fit, informed by Vapnik-Chervonenkis theory \cite{ref9,ref10}.

Dual Formulation: Leveraging Lagrange multipliers $\alpha_i \geq 0$, the primal dualizes to
\begin{align}
	\max_{\alpha} \quad & \sum_{i=1}^n \alpha_i - \frac{1}{2} \sum_{i,j=1}^n \alpha_i \alpha_j y_i y_j (\mathbf{x}_i \cdot \mathbf{x}_j) \\
	\text{s.t.} \quad & \sum_{i=1}^n \alpha_i y_i = 0,\nonumber\\
	\quad &0 \leq \alpha_i \leq C, \forall i = 1, \dots, n.\nonumber
\end{align}
The solution recovers $\mathbf{w} = \sum_{i=1}^{n} \alpha_i y_i \mathbf{x}_i$, emphasizing the role of support vectors where $\alpha_i > 0$. This form facilitates kernel extensions for nonlinearity \cite{ref9,ref10}.
\subsection{ Nonlinear Extensions }
Real-world data frequently defies linear separability in the ambient space. SVMs circumvent this via the kernel trick, implicitly mapping inputs to a higher-dimensional feature space $\mathcal{H}$ through a nonlinear transformation $\phi: \mathbb{R}^d \to \mathcal{H}$, where linear separation becomes feasible. The dual objective generalizes by substituting inner products with kernel functions $K(\mathbf{x}_i, \mathbf{x}_j) = \phi(\mathbf{x}_i) \cdot \phi(\mathbf{x}_j)$, yielding
\begin{align}
	\max_{\alpha} \quad & \sum_{i=1}^n \alpha_i - \frac{1}{2} \sum_{i,j=1}^n \alpha_i \alpha_j y_i y_j K(\mathbf{x}_i, \mathbf{x}_j) \\
	\text{s.t.} \quad & \sum_{i=1}^n \alpha_i y_i = 0,\nonumber\\
	\quad &0 \leq \alpha_i \leq C, \forall i = 1, \dots, n.\nonumber
\end{align}
 Mercer's theorem ensures $K$ corresponds to a valid $\phi$ if positive semi-definite, enabling computation without explicit $\phi$ (e.g., radial basis function $K(\mathbf{x}, \mathbf{z}) = \exp(-\gamma \|\mathbf{x} - \mathbf{z}\|^2)$, polynomial kernels). This duality underpins SVMs' flexibility, though kernel selection and hyperparameter tuning (e.g., via cross-validation) remain critical \cite{ref9}.\\
In summary, SVMs' elegance lies in their margin maximization paradigm, convex solvability, and kernel-induced expressivity, rendering them a benchmark for interpretable, theoretically sound learning, particularly in scenarios with structured sparsity or moderate data volumes. Subsequent sections explore SVM integrations within multi-task frameworks, building on these foundations.

\section{Twin Support Vector Machines: A Primer}

Twin Support Vector Machines constitute a significant evolution of classical Support Vector Machines, tailored for binary classification tasks. Pioneered by Jayadeva, Khemchandani, and Chandra \cite{ref11}, TWSVMs enhance computational efficiency and generalization by deriving two non-parallel hyperplanes, each optimized to be proximal to instances of one class while distal to the other. This dual-hyperplane strategy contrasts with the single hyperplane in standard SVMs, resulting in the resolution of two smaller quadratic programming problems (QPPs) rather than one large one, thereby reducing complexity and improving scalability. TWSVMs have undergone extensive scrutiny and extensions since their inception, exhibiting superior performance in domains such as pattern recognition, bioinformatics, and financial forecasting \cite{ref75, ref76, ref77}. This primer offers a thorough overview of TWSVMs, encompassing their geometric and algebraic formulations in linear and nonlinear contexts (including primal and dual representations), kernel extensions, duality principles, regularization techniques, and a comparative analysis with classical SVMs. While TWSVMs inherit SVMs' theoretical robustness, their class-specific modeling renders them particularly adept for imbalanced or noisy data sets.
\subsection{Linear TWSVM Formulation}
\begin{figure}
	\centering
	\includegraphics[scale=0.4]{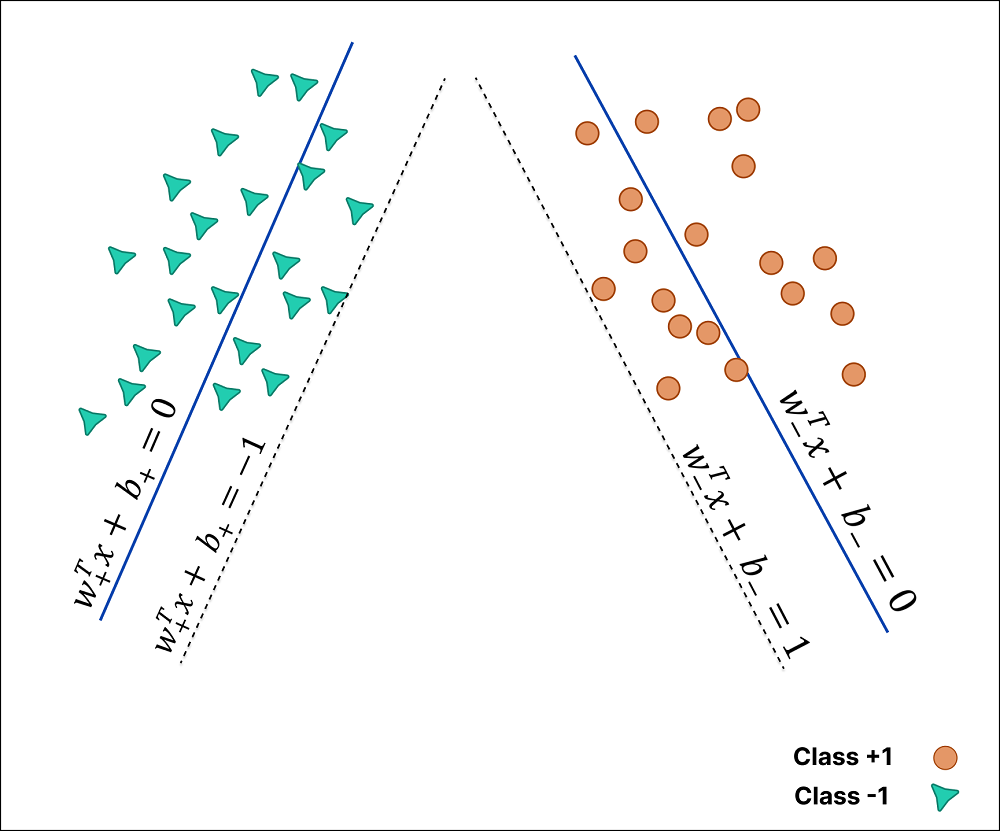}
	\caption{Geometric illustration of the TWSVM framework.}
	\label{fig:TSVM}
\end{figure}

The geometric foundation of TWSVMs revolves around constructing two non-parallel hyperplanes, each tailored to one class in a binary classification setting. Consider a data set comprising samples from two classes: the positive class represented by matrix $A \in \mathbb{R}^{m_1 \times d}$ (with $m_1$ denoting the number of positive samples and $d$ the feature dimension) and the negative class by matrix $B \in \mathbb{R}^{m_2 \times d}$. The hyperplanes are parameterized as $\mathbf{x}^\top \mathbf{w}_+ + b_+ = 0$ and $\mathbf{x}^\top \mathbf{w}_- + b_- = 0$, where $\mathbf{w}_+, \mathbf{w}_- \in \mathbb{R}^d$ are the normal vectors, and $b_+, b_- \in \mathbb{R}$ are the bias terms.
In the linear case, TWSVMs address two QPPs to minimize the squared distance to the points of one class while enforcing that points of the opposing class lie at least a unit distance away, incorporating slack variables for margin violations. For the positive class hyperplane, the primal optimization is:
\begin{align}
	\min_{\mathbf{w}_+, b_+, \xi_-} \quad & \frac{1}{2} \| A \mathbf{w}_+ + b_+ e_1 \|^2 + c_1 e_2^\top \xi_- \\
	\text{s.t.} \quad & -(B \mathbf{w}_+ + b_+ e_2) + \xi_- \geq e_2, \quad \xi_- \geq 0,\nonumber
\end{align}

where $e_1 \in \mathbb{R}^{m_1}$ and $e_2 \in \mathbb{R}^{m_2}$ are vectors of ones, $\xi_- \in \mathbb{R}^{m_2}$ are slack variables accommodating violations, and $c_1 > 0$ is a regularization parameter balancing structural and empirical risk. Symmetrically, for the negative class:
\begin{align}
	\min_{\mathbf{w}_-, b_-, \xi_+} \quad & \frac{1}{2} \| B \mathbf{w}_- + b_- e_2 \|^2 + c_2 e_1^\top \xi_+ \\
	\text{s.t.} \quad & (A \mathbf{w}_- + b_- e_1) + \xi_+ \geq e_1, \quad \xi_+ \geq 0,\nonumber
\end{align}

with $c_2 > 0$ and $\xi_+ \in \mathbb{R}^{m_1}$. This class-specific approach fosters enhanced generalization, particularly in imbalanced data sets, by allowing independent tuning of penalties for each class. Figure \ref{fig:TSVM} illustrates the classification of two classes using TWSVM.
The dual formulations enable efficient optimization and highlight support vectors. For the positive class primal, introducing Lagrange multipliers $\alpha \in \mathbb{R}^{m_2}$ yields the dual:
\begin{align}
	\max_{\alpha} \quad &  e_2^\top \alpha - \frac{1}{2} \alpha^\top G (F^\top F + \frac{1}{c_1} I)^{-1} G^\top \alpha \\
	\text{s.t.} \quad & 0 \leq \alpha \leq c_1 e_2,\nonumber
\end{align}

where $F = [A \ e_1]$, $G = [B \ e_2]$, and $I$ is the identity matrix. A parallel dual exists for the negative class. These duals, solvable via methods like SMO, embody sparsity: non-zero $\alpha$ (or $\boldsymbol{\beta}$ for the negative dual) correspond to support vectors, which uniquely determine the hyperplanes. The decision function assigns a new point $x$ to the class whose hyperplane is closer, i.e., $\arg\min_{k=\pm } \frac{|x^\top \mathbf{w}_k + b_k|}{\|\mathbf{w}_k\|}$.
\subsection{ Nonlinear Extensions }
For nonlinearly separable data, TWSVMs leverage the kernel trick, implicitly mapping inputs to a higher-dimensional Hilbert space $\mathcal{H}$ via a feature map $\phi: \mathbb{R}^d \to \mathcal{H}$. The hyperplanes transform to $K(\mathbf{x}^\top, C^\top) \mathbf{u}_+ + b_+ = 0$ and $K(\mathbf{x}^\top, C^\top) \mathbf{u}_- + b_- = 0$, where $K$ is a kernel function, and $C^\top = [A^\top \ B^\top]$ concatenates the data matrices.
The nonlinear primal for the positive class is:
\begin{align}
	\min_{\mathbf{u}_+, b_+, \xi_-} \quad &  \frac{1}{2} \| K(A, C^\top) \mathbf{u}_+ + b_+ e_1 \|^2 + c_1 e_2^\top \xi_-\\
	\text{s.t.} \quad &-(K(B, C^\top) \mathbf{u}_+ + b_+ e_2) + \xi_- \geq e_2, \quad \xi_- \geq 0.\nonumber
\end{align}
and
\begin{align}
	\min_{\mathbf{u}_-, b_-, \xi_+} \quad &  \frac{1}{2} \| K(B, C^\top) \mathbf{u}_- + b_- e_2 \|^2 + c_1 e_1^\top \xi_+\\
	\text{s.t.} \quad &(K(A, C^\top) \mathbf{u}_- + b_- e_1) + \xi_+ \geq e_1, \quad \xi_+ \geq 0.\nonumber
\end{align}

The kernelized dual for the positive class becomes:
\begin{align}
	\max_{\alpha} \quad & e_2^\top \alpha - \frac{1}{2} \alpha^\top K(B, C^\top) (K(A, C^\top)^\top K(A, C^\top) + \frac{1}{c_1} I)^{-1} K(B, C^\top)^\top \alpha\\
	\text{s.t.} \quad &0 \leq \alpha \leq c_1 e_2.\nonumber
\end{align}

The negative class dual mirrors this structure. Kernel parameters (e.g., $\gamma$) introduce additional regularization, enabling TWSVMs to capture complex, nonlinear relationships without explicit computation of $\phi$, while preserving sparsity and efficiency. Solutions recover $\mathbf{u}_+$ and $b_+$ (and analogs) via matrix inversions, with support vectors defined by non-zero multipliers.

\section{Comparison between Standard SVM and TWSVM}
In contrast to the standard SVM, which solves a single QPP to maximize the margin between the convex hulls of two classes, the TWSVM employs two smaller QPPs, each tailored to one class with constraints influenced by the opposing class. The standard SVM formulation results in a larger QPP involving all $m$ data points simultaneously, whereas TWSVM's dual-problem structure processes approximately $m/2$ points per QPP for a data set of size $m$. Consequently, TWSVM solves two compact QPPs, making it significantly faster than traditional SVM. Specifically, TWSVM achieves approximately a fourfold speedup, as it addresses two problems of size $m/2$, each with a complexity of $O((m/2)^3) = O(m^3/8)$, yielding a total complexity of $O(2 \times m^3/8) = O(m^3/4)$, compared to SVM's $O(m^3)$.

This efficiency gain in TWSVM does not compromise classification accuracy. Its dual non-parallel hyperplanes, each proximal to one class, enable more focused separation, particularly effective for imbalanced or non-linearly separable data, unlike SVM’s single hyperplane. For example, TWSVM excels in applications like image classification or bioinformatics, where class distributions may be uneven. However, TWSVM requires careful tuning of its regularization parameters $c_1$ and $c_2$, akin to the $C$ parameter in SVM, to optimize performance.

To delineate the distinctions between TWSVM and classical SVM, Table~\ref{tab:svm_vs_twsvm} provides a comparative analysis across key dimensions, including formulation, geometric interpretation, computational complexity, scalability, generalization, advantages/limitations, and task adaptability. This comparison underscores TWSVM's efficiency and flexibility, particularly in scenarios demanding rapid training and adaptability to complex class distributions, such as multi-task learning frameworks.

\begin{table}
	\centering
	\caption{Advantages and Challenges of Classical SVM and TWSVM}
	\label{tab:svm_vs_twsvm}
	\begin{tabular}{l p{5cm} p{5cm}}
		\toprule
		\textbf{Aspect} & \textbf{Classical SVM} & \textbf{TWSVM} \\
		\midrule
		\textbf{Formulation} & Single QPP with $m = m_1 + m_2$ constraints to maximize a single margin hyperplane; uses regularization parameter $C$ to balance margin and error. & Two smaller QPPs (each with $\approx m/2$ constraints) for non-parallel hyperplanes, using $c_1$, $c_2$ to optimize class-specific proximity and separation. \\
		\textbf{Geometric Property} & One hyperplane maximizes equidistant margin ($2/\|w\|$); effective for balanced, linearly separable data. & Two non-parallel hyperplanes, each proximal to its class and distal to the other; adaptable to imbalanced or complex distributions. \\
		\textbf{Computational Cost} & $O(m^3)$ for a single large QPP; mitigated by SMO for large data sets. & $O(m^3/4)$ total for two QPPs, each $O((m/2)^3)$, offering up to 4x faster training; enhanced by variants like LS-TWSVM (Least Squares TWSVM). \\
		\textbf{Scalability} & Limited for large data sets; relies on optimization techniques for efficiency. & Improved scalability with smaller problems; suitable for big data with linear system solvers. \\
		\textbf{Generalization} & Strong via margin maximization and VC theory; may falter with imbalance or noise. & Enhanced in imbalanced or correlated tasks due to class-specific modeling, reducing overfitting in complex cases. \\
		\textbf{Advantages} & Robust theoretical basis, extensive multi-class support (e.g., one-vs-one); excels in high-dimensional settings. & Faster training, better imbalance handling; flexible with many variants. \\
		\textbf{Limitations} & High computational load for large $m$; sensitive to imbalance without tuning. & Less mature multi-class support; dual parameter tuning ($c_1$, $c_2$) adds complexity. \\
		\textbf{Task Adaptability} & Uniform margin may underperform with heterogeneous tasks. & Class-specific hyperplanes improve performance across diverse task distributions, e.g., multi-task learning. \\
		\bottomrule
	\end{tabular}
\end{table}

The comparison in Table~\ref{tab:svm_vs_twsvm} highlights TWSVM's computational efficiency and adaptability, making it a promising approach for applications requiring rapid training and robust handling of complex or imbalanced data sets, as further explored in multi-task learning contexts.

\section{Multi-Task Learning Based on SVM}

Support Vector Machines have been a cornerstone of machine learning due to their robust theoretical foundations and effectiveness in high-dimensional and data-scarce scenarios. Extending SVMs to MTL leverages their margin-maximization principle to exploit task relatedness, improving generalization across multiple related tasks. This section provides a comprehensive overview of MTL frameworks based on SVMs, starting with the foundational formulation, followed by recent extensions, representative models, optimization strategies, and empirical applications. The discussion aligns with the notation and structure established in previous sections, ensuring consistency with the chapter's scope.

\subsection{Formulation of Multi-Task Learning Based on SVM}

The foundational formulation for multi-task SVM (M-SVM) extends the classical SVM by introducing shared parameters or constraints to couple multiple tasks, promoting knowledge transfer while preserving task-specific characteristics \cite{ref6,ref12}. Following the notation from Section 2.1, consider $T$ supervised tasks, each with a data set
\[
\mathcal{D}_t = \{(\mathbf{x}_{ti}, y_{ti})\}_{i=1}^{n_t},
\]
where $\mathbf{x}_{ti} \in \mathbb{R}^d$, $y_{ti} \in \{-1, +1\}$, and $n_t$ is the number of samples for task $t$. The goal is to learn task-specific hyperplanes
\[
f_t(\mathbf{x}) = \mathbf{w}_t^\top \mathbf{x} + b_t, \quad \mathbf{w}_t \in \mathbb{R}^d, \quad b_t \in \mathbb{R},
\]
while enforcing relatedness across tasks.

A standard approach, as proposed by \cite{ref6}, assumes that each task's weight vector $\mathbf{w}_t$ is composed of a shared component $\mathbf{w}_0$ and a task-specific deviation $v_{t}$, i.e.,
\[
\mathbf{w}_t = \mathbf{w}_0 + v_{t}.
\]
This decomposition encourages tasks to have similar decision boundaries while allowing flexibility for task-specific patterns.

\begin{figure}
	\centering
	\includegraphics[scale=0.4]{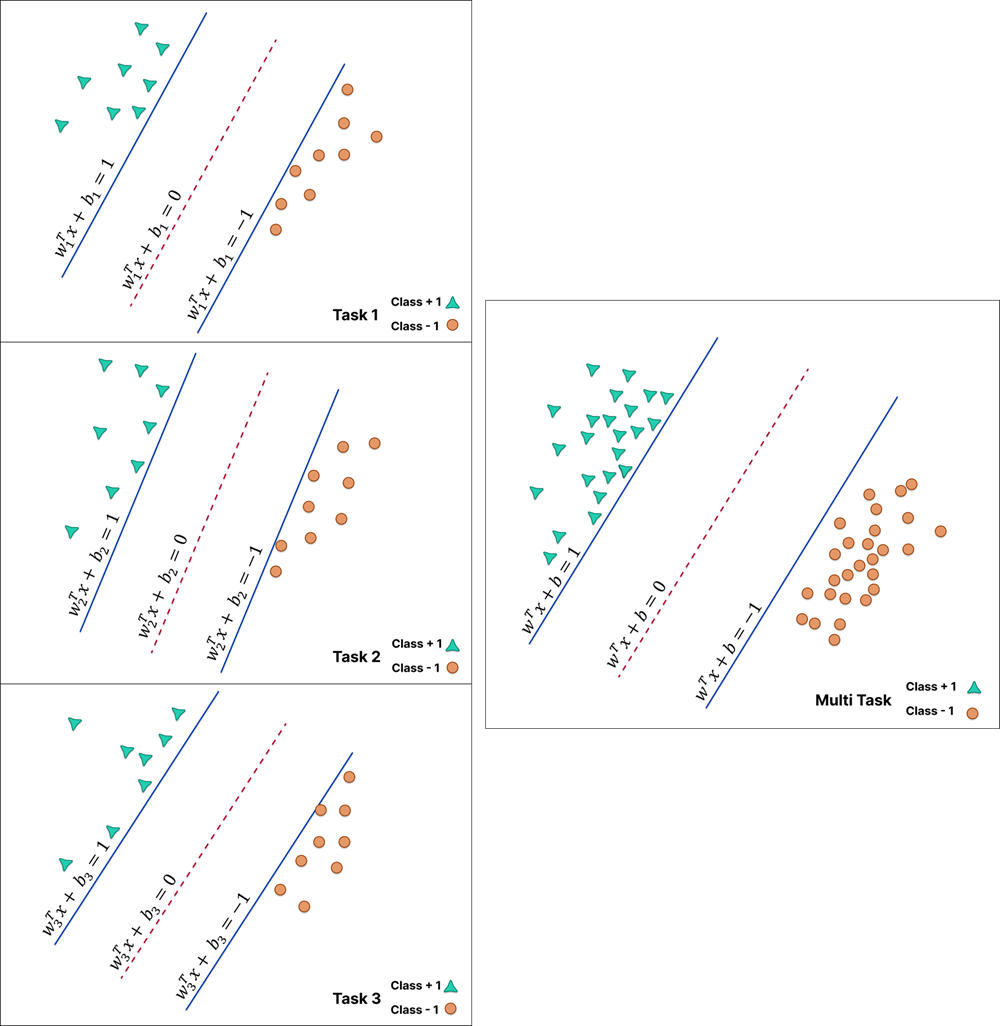}
	\caption{Geometric illustration of the M-SVM framework.}
	\label{fig:MTSVM}
\end{figure}

The primal optimization problem for linear M-SVM can be formulated as:
\begin{align}
	\min_{\mathbf{w}_0, \{v_{t}\}, \{b_t\}, \{\xi_t\}} \quad & 
	\frac{1}{2} \|\mathbf{w}_0\|^2 + \frac{\mu}{2} \sum_{t=1}^T \|v_{t}\|^2 + C \sum_{t=1}^T \sum_{i=1}^{n_t} \xi_{ti}, \\
	\text{s.t.} \quad & 
	y_{ti} ((\mathbf{w}_0 + v_{t})^\top \mathbf{x}_{ti} + b_t) \geq 1 - \xi_{ti}, \nonumber \\
	& \xi_{ti} \geq 0, \quad \forall i, \dots, n_t, \quad t = 1, \dots, T, \nonumber
\end{align}
where $\mu > 0$ controls the trade-off between the shared component $\mathbf{w}_0$ and task-specific deviations $v_{t}$, $C > 0$ is the penalty parameter for hinge loss, and $\xi_t = [\xi_{t1}, \dots, \xi_{tn_t}]^\top$ are slack variables. The term $\frac{1}{2} \|\mathbf{w}_0\|^2$ promotes a large margin for the shared hyperplane, while $\frac{\mu}{2} \sum_{t=1}^T \|v_{t}\|^2$ ensures task-specific deviations remain small, enforcing similarity across tasks. Figure \ref{fig:MTSVM} illustrates the classification of two classes using multi-task learning via SVM.

To derive the dual, we introduce Lagrange multipliers $\alpha_{ti} \ge 0$ for the margin constraints and $\mu_{ti} \ge 0$ for the slack variable constraints $\xi_{ti} \ge 0$. The Lagrangian function is:
\begin{align}
	\mathcal{L}(\mathbf{w}_0, \{v_t\}, \{b_t\}, \{\xi_t\}, \{\alpha_t\}, \{\mu_t\}) 
	&= \frac{1}{2} \|\mathbf{w}_0\|^2 + \frac{\mu}{2} \sum_{t=1}^T \|v_t\|^2 + C \sum_{t,i} \xi_{ti} \nonumber \\
	&\quad - \sum_{t=1}^T \sum_{i=1}^{n_t} \alpha_{ti} \big[y_{ti} ((\mathbf{w}_0 + v_t)^\top \mathbf{x}_{ti} + b_t) - 1 + \xi_{ti} \big] \nonumber \\
	&\quad - \sum_{t=1}^T \sum_{i=1}^{n_t} \mu_{ti} \xi_{ti}.
\end{align}

\paragraph{Step 1: Derivatives w.r.t. primal variables}
\begin{align}
	\frac{\partial \mathcal{L}}{\partial \mathbf{w}_0} &= \mathbf{w}_0 - \sum_{t=1}^T \sum_{i=1}^{n_t} \alpha_{ti} y_{ti} \mathbf{x}_{ti} = 0 
	\quad \Rightarrow \quad 
	\mathbf{w}_0 = \sum_{t=1}^T \sum_{i=1}^{n_t} \alpha_{ti} y_{ti} \mathbf{x}_{ti}, \\
	\frac{\partial \mathcal{L}}{\partial v_t} &= \mu v_t - \sum_{i=1}^{n_t} \alpha_{ti} y_{ti} \mathbf{x}_{ti} = 0 
	\quad \Rightarrow \quad 
	v_t = \frac{1}{\mu} \sum_{i=1}^{n_t} \alpha_{ti} y_{ti} \mathbf{x}_{ti}, \\
	\frac{\partial \mathcal{L}}{\partial b_t} &= - \sum_{i=1}^{n_t} \alpha_{ti} y_{ti} = 0 
	\quad \Rightarrow \quad 
	\sum_{i=1}^{n_t} \alpha_{ti} y_{ti} = 0, \\
	\frac{\partial \mathcal{L}}{\partial \xi_{ti}} &= C - \alpha_{ti} - \mu_{ti} = 0 
	\quad \Rightarrow \quad 0 \le \alpha_{ti} \le C.
\end{align}

\paragraph{Step 2: Dual problem}

Substituting the primal solutions back into the Lagrangian, the dual optimization problem becomes:
\begin{align}
	\max_{\{\alpha_{ti}\}} \quad & 
	\sum_{t=1}^T \sum_{i=1}^{n_t} \alpha_{ti} - \frac{1}{2} \sum_{t=1}^T \sum_{i,j=1}^{n_t} \alpha_{ti} \alpha_{tj} y_{ti} y_{tj} \mathbf{x}_{ti}^\top \mathbf{x}_{tj} \nonumber \\
	& - \frac{1}{2T} \sum_{t,s=1}^{T} \sum_{i=1}^{n_t} \sum_{j=1}^{n_s} \alpha_{ti} \alpha_{sj} y_{ti} y_{sj} \mathbf{x}_{ti}^\top \mathbf{x}_{sj}, \nonumber\\
	\text{s.t.} \quad & 0 \le \alpha_{ti} \le C, \quad \sum_{i=1}^{n_t} \alpha_{ti} y_{ti} = 0, \quad \forall t.
\end{align}

\paragraph{Step 3: Recovering primal solutions.}

The shared and task-specific weight vectors are reconstructed from the dual variables:
\begin{align}
	\mathbf{w}_0 &= \frac{1}{T} \sum_{t=1}^T \sum_{i=1}^{n_t} \alpha_{ti} y_{ti} \mathbf{x}_{ti}, \\
	v_t &= \sum_{i=1}^{n_t} \alpha_{ti} y_{ti} \mathbf{x}_{ti} - \mathbf{w}_0.
\end{align}

This derivation explicitly demonstrates how the dual formulation arises from the Lagrangian and elimination of primal variables, while respecting the parameter-sharing structure across multiple tasks.
Meanwhile, the label of a new sample $x$ in the $t^{th}$ task can be determined by:
\[
f_t(x) = \operatorname{sgn}[(w_0 + v_t) \cdot x]
\]
If $f_t(x) < 0$, then the label of sample $x$ is predicted to be negative. 
If $f_t(x) > 0$, the label of sample $x$ is predicted to be positive.

In many real-world applications, the data for each task may not be linearly separable in the original input space $\mathbb{R}^d$. To handle such cases, nonlinear M-SVM projects the input vectors into a high-dimensional feature space via a nonlinear mapping:
\[
\phi: \mathbb{R}^d \rightarrow \mathbb{R}^\ell, \quad \ell \gg d.
\]
This transformation allows linearly inseparable data in the original space to become linearly separable in the feature space. Consequently, the decision function for task $t$ is expressed as
\[
f_t(\mathbf{x}) = (\mathbf{w}_0 + v_t)^\top \phi(\mathbf{x}) + b_t,
\]
where $\mathbf{w}_0$ is the shared component across tasks, $v_t$ is the task-specific component, and $b_t$ is the bias term for task $t$.


For each training sample $(\mathbf{x}_{ti}, y_{ti})$, the classification constraint in the feature space becomes:
\[
y_{ti} \big( (\mathbf{w}_0 + v_t)^\top \phi(\mathbf{x}_{ti}) + b_t \big) \ge 1 - \xi_{ti}, \quad \xi_{ti} \ge 0,
\]
where $\xi_{ti}$ are slack variables that allow for soft-margin classification in case of overlapping classes.

Direct computation of $\phi(\mathbf{x})$ in high-dimensional space can be computationally expensive. The kernel trick circumvents this by computing inner products in the feature space implicitly using a kernel function:
\[
K(\mathbf{x}, \mathbf{x}') = \langle \phi(\mathbf{x}), \phi(\mathbf{x}') \rangle.
\]
This allows the optimization problem to be expressed entirely in terms of kernel evaluations without explicitly calculating $\phi(\mathbf{x})$.


	
	

Hence,  the primal and dual optimization problems are formulated in terms of $K(\mathbf{x}_{ti}, \mathbf{x}_{tj})$. In particular, the dual formulation for task $t$ becomes:
\begin{align}
	\max_{\alpha_t} \quad & 
	\sum_{i=1}^{n_t} \alpha_{ti} - \frac{1}{2} \sum_{i,j=1}^{n_t} \alpha_{ti} \alpha_{tj} y_{ti} y_{tj} K(\mathbf{x}_{ti}, \mathbf{x}_{tj})\nonumber \\
	&- \frac{1}{2T}\sum_{t,s=1}^{T} \sum_{i=1}^{n_t} \sum_{j=1}^{n_s} \alpha_{ti} \alpha_{sj} y_{ti} y_{sj} K(\mathbf{x}_{ti}^\top \mathbf{x}_{sj}), \nonumber\\ 
	\text{s.t.} \quad & 0 \le \alpha_{ti} \le C, \quad \sum_{i=1}^{n_t} \alpha_{ti} y_{ti} = 0.
\end{align}

The shared component $\mathbf{w}_0$ and task-specific components $v_t$ are then recovered from the support vectors, ensuring both knowledge transfer and task-specific adaptability.

Geometrically, the mapping $\phi(\cdot)$ lifts the input data into a high-dimensional space where a hyperplane can separate the classes. The shared component $\mathbf{w}_0$ represents a global orientation of the hyperplane influenced by all tasks, while $v_t$ adjusts the hyperplane to account for task-specific variations. The combination ensures that tasks benefit from shared information while maintaining flexibility for individual differences.


\begin{remark}
	Single-task SVM trains one independent classifier per task, yielding overall training cost 
	$O(T m^{3})$ when each of the $T$ tasks has $m$ training examples 
	(assuming an $O(n^{3})$ solver for $n$ examples). 
	The regularized multi-task formulation of Evgeniou and Pontil jointly optimizes across all tasks 
	by treating the pooled data set of size $Tm$ as a single SVM problem \cite{ref6}; 
	naively this implies a training cost of 
	\[
	O\big((Tm)^{3}\big) = O(T^{3} m^{3}).
	\]
	The multi-task approach exchanges statistical information across tasks 
	(via shared regularization and task-specific components), 
	which can substantially improve generalization when tasks are related 
	or per-task data are scarce. 
	However, this gain comes at a potentially large computational cost 
	for moderate-to-large $T$ or $m$.	
\end{remark}
\subsection{Recent Extensions of SVM for Multi-Task Learning}
In this section, we provide a comprehensive review of recent advancements in MTL models based on SVM. This narrative traces the evolution from foundational models to sophisticated extensions, emphasizing how each innovation addresses key limitations such as computational complexity, noise sensitivity, outlier robustness, class imbalance, and task heterogeneity.

As one of the earliest contributions to integrating MTL with SVMs, Evgeniou and Pontil  introduced the Multi-task Support Vector Machine (M-SVM) \cite{ref6}, a pioneering framework that leverages inter-task correlations to enhance predictive performance. By incorporating shared hyperplane parameters alongside task-specific deviations, M-SVM enables joint learning of related tasks through a single QPP, solved using Lagrange duals and Sequential Minimal Optimization. This approach marked a significant advancement in capturing task relationships.

Building on early advancements in multi-task learning, Cai and Cherkassky introduced SVM$+$MTL regression \cite{ref12}, a novel method that extends the SVM$ +$ \cite{ref22} framework to regression by incorporating MTL principles. This approach addresses the shortcomings of traditional SVM regression, which fails to leverage group-specific structures in data, by treating data groups as tasks to capture their interrelationships, thereby improving generalization through the Learning with Structured Data (LWSD) framework. To tackle the challenge of managing multiple tuning parameters, the authors propose a hybrid model selection strategy that integrates analytical optimization for standard SVM regression with resampling-based techniques for SVM$+$  and SVM$+$MTL parameters. 

Xu et al. in 2014 addressed the limitation of traditional single-task learning methods, which overlook the inherent interactions between multiple tasks \cite{ref13}. They proposed a novel multi-task learning approach, termed Multi-Task Least Squares Support Vector Machine (MTLS-SVM), extending the Least Squares Support Vector Machine (LS-SVM) framework \cite{ref23} by drawing inspiration from Regularized Multi-Task Learning (RMTL) \cite{ref6}. They demonstrate that LS-SVM's generalization performance is comparable to standard SVM. The proposed MTLS-SVM unifies classification and regression tasks within an efficient training algorithm that leverages Krylov methods, solving only a convex linear system during training, similar to LS-SVM. 

In 2015, in response to the challenge of single-task learning's inability to leverage relationships among related tasks, particularly in the context of the growing importance of visual recognition due to the widespread use of imagery, the authors of \cite{ref14} proposed a novel multi-task learning model based on the proximal support vector machine (PSVM) \cite{ref25}. Unlike single-task learning, which trains each task independently, their multi-task PSVM (MT-PSVM) jointly learns multiple tasks by exploiting inter-task relationships, thereby enhancing performance. Building on the PSVM framework, which adopts the large-margin concept of standard support vector machines but with relaxed constraints and lower computational cost, the proposed MT-PSVM retains these advantages while outperforming other popular multi-task learning models. 

In 2016, to counter the issue of ineffective feature selection in scenarios with limited labeled data, where traditional single-task learning methods fail to exploit shared information across related tasks, \cite{ref15} propose a novel multi-task support vector machine (SVM)-based feature selection method. Their approach simultaneously learns discriminative features for multiple tasks by enforcing sparsity in the feature selection matrix for each task using an l1-norm penalty, ensuring only the most relevant features are selected. Additionally, they incorporate a global low-rank constraint on the combined feature selection matrices to capture shared structures across tasks, enhancing performance by leveraging inter-task relatedness. Instead of directly minimizing the non-convex rank of the matrix, they optimize the convex trace norm, using a gradient descent-based convex optimization method to achieve a global optimum. 

To resolve the challenges of sensitivity to noise features in MTLS-SVM \cite{ref13}, which limits its effectiveness despite its ability to leverage shared information across tasks for improved learning performance, in 2020, the authors of \cite{ref17} proposed a novel algorithm called Sparse MTLS-SVM (SMTLS-SVM). This approach extends MTLS-SVM by introducing a 0–1 multi-task feature vector to control feature selection, enabling the model to learn a sparse subset of informative features for classification. SMTLS-SVM employs an efficient alternating minimization method to iteratively select the most relevant features, enhancing robustness against noise. 

In 2021, to overcome the problem of underutilizing non-target task data in MTL, which typically focuses only on target task data and overlooks valuable prior knowledge from non-target tasks, the researchers proposed a novel Multi-Task Support Vector Machine with Universum Data (U-MTLSVM) \cite{ref19}. This approach enhances MTL by incorporating Universum data that data not belonging to any target task categories as prior knowledge to improve classifier performance. By assigning corresponding Universum data to each task, the method leverages the inherent relationships among tasks to enrich the learning process. The optimization problem is solved using the Lagrange method to derive multi-task classifiers.

To handle the challenges of noise sensitivity and high computational cost in traditional SVM and Multi-Task Learning Support Vector Machines, which rely on non-differentiable hinge or pinball loss functions and require solving complex QPP, in 2022, \cite{ref21} et al proposed a novel Multi-Task Support Vector Machine with Generalized Huber Loss (GHMTSVM). This approach extends the Generalized Huber SVM (GHSVM) \cite{ref26} to multi-task learning by integrating the differentiable and noise-robust generalized huber loss, inspired by RMTL. GHMTSVM leverages shared representations across multiple tasks to enhance generalization, using functional iteration to find optimal solutions without solving QPP, thus improving computational efficiency.

In 2023, to address the limitations of single-task learning, which fails to capture relationships among related tasks, \cite{ref27} et al proposed two novel multi-task learning algorithms, the Multi-Task Learning Asymmetric Proximal Support Vector Machine (MTL-a-PSVM) and its extension (EMTL-a-PSVM), designed for classification tasks. These algorithms overcome the shortcomings of traditional proximal support vector machines by replacing the standard squared loss function with an asymmetric squared loss function, which enhances generalization performance through adjustable asymmetric parameters. By leveraging two multi-task assumptions about task relatedness, the proposed methods simultaneously train multiple tasks, solving corresponding quadratic programming problems to achieve optimal solutions. 

To address the challenges of insufficient sample utilization and class imbalance in multi-task, multi-class classification, Xu et al. \cite{ref28} proposed in 2024 a novel Multi-Task K-Class Support Vector Classification and Regression (MTKSVCR) model. This approach combines the strengths of RMTL \cite{ref6} and the KSVCR framework \cite{ref29}, employing a "one-versus-one-versus-rest" strategy to leverage all class samples simultaneously, thereby enhancing testing accuracy. The model decomposes weight vectors into task-common and task-specific components, effectively mining shared knowledge across tasks by adjusting penalty parameters.

In 2025, to overcome the challenges of poor generalization and high computational cost in traditional SVMs and MTL-SVMs that rely on non-differentiable hinge loss and complex QPP, Liu et al. \cite{ref30} proposed a novel Multi-Task Learning Generalized Huber Loss Support Vector Machine (MTL-GHSVM) for binary classification, covering both linear and non-linear cases. Inspired by GHSVM \cite{ref26} and RMTL \cite{ref6}, this method extends GHSVM to multi-task settings by incorporating the differentiable generalized Huber loss, which enhances robustness and generalization compared to hinge loss-based SVMs and GHSVM. 

These extensions enhance MTSVM’s flexibility, addressing challenges like noise, task heterogeneity, sparsity, and scalability, making them suitable for diverse applications such as multimedia analysis, renewable energy forecasting, and medical diagnostics.

To provide a structured overview of the evolution of MTL models based on SVMs, we summarize representative contributions in Table \ref{tab:mtl-svm}. This table highlights the progression of SVM-based MTL approaches across different years, emphasizing their key features and the optimization strategies employed. The models range from early formulations introducing task-coupling parameters and kernel adaptations, to more advanced variants that integrate Universum data, robust loss functions, and proximal optimization techniques for improved generalization and scalability.
\begin{longtable}{l l p{5.5cm} p{5.5cm}}
	\caption{Representative Models in Multi-Task Learning based on SVM} \label{tab:mtl-svm} \\
	\hline
	\textbf{Model} & \textbf{Year} & \textbf{Key Features} & \textbf{Optimization Approach} \\
	\hline
	\endfirsthead
	
	\multicolumn{4}{c}{{\bfseries Table \thetable\ (continued)}} \\
	\hline
	\textbf{Model} & \textbf{Year} & \textbf{Key Features} & \textbf{Optimization Approach} \\
	\hline
	\endhead
	
	\hline \multicolumn{4}{r}{{Continued on next page}} \\
	\endfoot
	
	\hline
	\endlastfoot
	
	\cite{ref6} & 2004 & Extends SVMs to multi-task learning using a task-coupling parameter ($\mu$) in a novel kernel function, supports simultaneous learning of related tasks, reduces to single-task learning when µ is large & Minimization of regularization functionals with a matrix-valued kernel, tuned by task-coupling parameter µ \\
	
	\cite{ref12} & 2008 & Adaptation of SVM+ for multi-task learning, uses two Hilbert spaces (decision and correcting) for each task, incorporates a common decision function shared across tasks and unique correcting functions, requires group labels for test data & Minimization of a regularized loss function with 2-norm penalties on decision and correcting functions, tuned by parameters $C$ and $\gamma$, using linear kernel for decision space and Gaussian kernel for correcting space \\
	
	\cite{ref13} & 2014 & Extends LS-SVM to multi-task learning, solves convex linear system for both classification and regression, uses matrix-valued kernel inspired by RMTL, employs Krylov methods for efficient training & Minimization of a convex linear system with regularization, tuned by parameters $\gamma$ and $\lambda$, using linear or RBF kernel functions \\
	
	\cite{ref14} & 2015 & Extends PSVM to multi-task learning, uses large-margin idea with looser constraints, supports both classification and regression, optimized for efficiency with low computational cost, effective for small training datasets & Solves a quadratic optimization problem with equality constraints, tuned by parameters $\nu$ and $\lambda$, using linear kernel for classification and regression tasks \\
	
	\cite{ref15} & 2016 & Uses hinge loss with $l_{1,2}$-norm for sparse feature selection per task, imposes low-rank constraint via trace norm to exploit shared knowledge across tasks, suitable for multimedia applications with limited labeled data & Convex optimization minimizing hinge loss with $l_{1,2}$-norm regularization and trace norm for low-rank constraint, solved via gradient descent \\
	
	\cite{ref17} & 2020 & Extends MTLS-SVM with 0–1 feature vector for sparse feature selection, iteratively selects informative features, robust to noise features, improves classification performance via MTL & Solves linear equations per task with alternating minimization, uses linear kernel \\
	
	\cite{ref19} & 2021 & Integrates Universum data into multi-task SVM, assigns task-specific Universum data as prior knowledge, uses shared parameter $\mu$ to link tasks, improves generalization by refining hyperplanes, suitable for text classification & Solves dual optimization problem using Lagrangian method, employs linear kernel for text data \\
	
	\cite{ref21} & 2022 & Extends GHSVM to multi-task learning, uses differentiable generalized huber loss for noise insensitivity and robustness, employs functional iterative method for faster computation, suitable for large-scale and noisy datasets & Convex optimization using functional iterative method to minimize generalized huber loss \\
	
	\cite{ref27} & 2023 & Multitask learning to capture relationships among tasks, replaces squared loss with asymmetric squared loss to improve generalization, proposes two multitask proximal SVM algorithms (MTL-a-PSVM, EMTL-a-PSVM) & Solves quadratic programming problem (QPP) with proximal SVM optimization under asymmetric loss \\
	
	\cite{ref28} & 2024 & Extends RMTL to multi-task multi-class via ``one-versus-one-versus-rest'' strategy, uses task-common and task-specific regularization for knowledge sharing, handles all samples simultaneously to avoid imbalance, employs multi-parameter safe acceleration (SA) rule for efficiency & Solves QPPs with dual optimization and variational inequality-based SA rule to screen redundant samples, tunes parameters via grid search, safe screening ensures identical solutions \\
	
	\cite{ref30} & 2025 & Extends GHSVM to multi-task learning, uses differentiable generalized Huber loss for better generalization, leverages task correlations, supports linear and Gaussian kernels, avoids QPP by using functional iteration & Solves via functional iterative methods, employs linear or Gaussian RBF kernel, uses 3-fold cross-validation \\
	
\end{longtable}

The corresponding evaluation metrics reported in the literature are consolidated in Table \ref{tab:mtl-svm-eval} that outlines the performance factors commonly used to benchmark MTL-SVM models, including RMSE, Explained Variance (EV), Accuracy, Average AUC, Running Time, Mean Average Precision (MAP), and statistical significance tests. 
\begin{table}
	\centering
	\caption{Evaluation Factors in Multi-Task Learning based on SVM} \label{tab:mtl-svm-eval}
	\begin{tabular}{|c|c|c|c|c|c|c|}
		\hline
		Model & RMSE & EV & Accuracy & Average AUC & Running Time & MAP \\
		\hline
		\cite{ref6}   & $\checkmark$ & $\checkmark $&             &             &             &             \\
		\cite{ref12}  &            &            & $\checkmark $ &             &             &             \\
		\cite{ref13} &            & $\checkmark $ &             &             &             &             \\
		\cite{ref14}   &            &$\checkmark $ &             &$\checkmark $ & $\checkmark $  &$\checkmark $\\
		\cite{ref15}  &            &            &             &             &             & $\checkmark $ \\
		\cite{ref17}  &            &            & $\checkmark $  &             &             &             \\
		\cite{ref19}   &            &            & $\checkmark $ &             & $\checkmark $  &             \\
		\cite{ref21}   &            &            &$\checkmark $ &             & $\checkmark $ &             \\
		\cite{ref27}   &            &            & $\checkmark $  &             &             &             \\
		\cite{ref28} &            &            & $\checkmark $ &             & $\checkmark $  &             \\
		\cite{ref30} &            & $\checkmark $ & $\checkmark $& $\checkmark $  &             &             \\
		\hline
	\end{tabular}
\end{table}

\section{Multi-Task Learning based on TWSVM}
While SVM-based MTL has a rich literature, the extension of  TWSVM to multi-task settings is a more recent and emerging area. TWSVM's inherent efficiency, achieved by solving two smaller QPPs for non-parallel hyperplanes, makes it particularly appealing for MTL, where scalability across multiple tasks is crucial. This section delves into MTL frameworks grounded in TWSVM, highlighting recent extensions that incorporate task relatedness through shared structures, robust loss functions, and prior knowledge. We first present the foundational formulation, followed by representative models, their optimization strategies, and empirical applications, drawing comparisons to underscore their advantages in generalization and computational efficiency.

\subsection{Formulation of Multi-Task Learning based on TWSVM}
The foundational formulation for multi-task TWSVM, known as the Direct Multi-Task Twin Support Vector Machine (DMTSVM), extends the single-task TWSVM by introducing shared hyperplane components across tasks to exploit inter-task correlations. This parameter-based sharing assumes that the hyperplanes for each class deviate from common mean hyperplanes in a controlled manner \cite{ref31}.

Recall the notation from subsection 1.2.1: $T$ tasks with data sets $\mathcal{D}_t = \{(\mathbf{x}_{ti}, y_{ti})\}_{i=1}^{n_t}$, $\mathbf{x}_{ti} \in \mathbb{R}^d$, $y_{ti} \in \{-1, +1\}$. For each task $t$, construct augmented matrices $A_t \in \mathbb{R}^{m_{1t} \times (d+1)}$ whose rows are $[\mathbf{x}_{ti}^\top, 1]$ for positive samples ($y_{ti} = +1$), and $B_t \in \mathbb{R}^{m_{2t} \times (d+1)}$ for negative samples ($y_{ti} = -1$). The global matrices are $A = \begin{bmatrix} A_1^\top,..., A_T^\top \end{bmatrix}^\top \in \mathbb{R}^{( \sum_{t=1}^{T} m_{1t} ) \times (d+1)}$ and $B = \begin{bmatrix} B_1^\top, ..., B_T^\top \end{bmatrix}^\top \in \mathbb{R}^{( \sum_{t=1}^{T} m_{2t} ) \times (d+1)}$.

The hyperplanes for the positive class in task $t$ are defined via shared $u_{0} = [\mathbf{w}_1^\top, b_1]^\top \in \mathbb{R}^{d+1}$ and task-specific $u_{t} = [\mathbf{w}_{1t}^\top, b_{1t}]^\top \in \mathbb{R}^{d+1}$, such that the effective hyperplane is $u_{0} + u_{t}$. Similarly for the negative class with $v_{0}$ and $v_{t}$.

\begin{figure}
	\centering
	\includegraphics[scale=0.6]{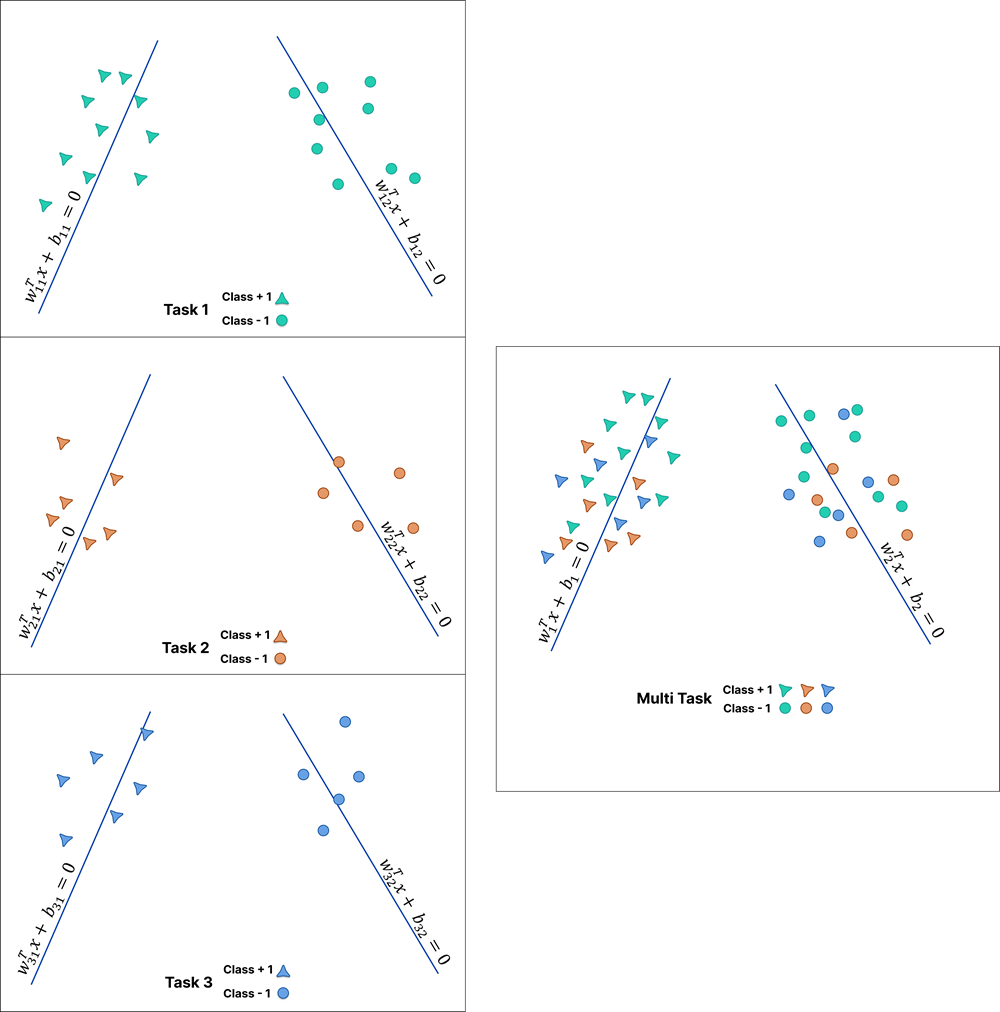}
	\caption{Geometric illustration of the DMTSVM framework.}
	\label{fig:DMTSVM}
\end{figure}

The primal optimization for the positive hyperplanes is:

\begin{align}
	\mathop {\min }_{u_{0}, u_{t}, \xi_t}\,  & \frac{1}{2} \| A u_{0} \|^2 + \frac{\mu_1}{2} \sum_{t=1}^T \| A_t u_{t} \|^2 + c_1 \sum_{t=1}^T e_{2t}^\top \xi_t,\nonumber \\
	\text{s.t.}\quad &-(B_t (u_{0} + u_{t})) + \xi_t \geq e_{2t}, \quad \xi_t \geq 0, \quad \forall t = 1, \dots, T,
\end{align}
and 
\begin{align}
	\mathop {\min }_{v_{0}, v_{t}, \eta_t}\,  & \frac{1}{2} \| B v_{0} \|^2 + \frac{\mu_2}{2} \sum_{t=1}^T \| B_t v_{t} \|^2 + c_2 \sum_{t=1}^T e_{1t}^\top \eta_t, \nonumber\\
	\text{s.t.}\quad &(A_t (v_{0} + v_{t})) + \eta_t \geq e_{1t}, \quad \eta_t \geq 0, \quad \forall t = 1, \dots, T,
\end{align}

where $\mu_1,\mu_2 > 0$ regulates task similarity (larger values enforce stronger sharing), $c_1,c_2 > 0$ are the penalty parameter, $e_{2t} \in \mathbb{R}^{m_{2t}}$ and  $e_{1t} \in \mathbb{R}^{m_{1t}}$ are the all-ones vector, and $\xi_t \in \mathbb{R}^{m_{2t}}$, $\eta_t \in \mathbb{R}^{m_{1t}}$ are slack variables. 
Figure~\ref{fig:DMTSVM} provides a geometric illustration of the DMTSVM framework. The method classifies the data into two categories, namely the $+1$ and $-1$ classes. Within each category, the data are further partitioned into three tasks, resulting in three distinct subgroups for both the $+1$ and $-1$ classes. Subsequently, for each task, three hyperplanes are constructed: one corresponding to the $+1$ class and another corresponding to the $-1$ class.

In the following, we have obtained the dual formulation of the primal DMTSVM problem for the positive class hyperplanes.
Hence, introduce Lagrange multipliers $\alpha_t \ge 0$ for the inequality constraints and $\beta_t \ge 0$ for the non-negativity of slack variables. The Lagrangian function is:
\begin{align}
	\mathcal{L}(u_0, \{u_t\}, \{\xi_t\}, \{\alpha_t\}, \{\beta_t\}) 
	&= \frac{1}{2} \| A u_0 \|^2 + \frac{\mu_1}{2} \sum_{t=1}^T \| A_t u_t \|^2 + c_1 \sum_{t=1}^T e_{2t}^\top \xi_t \nonumber \\
	&\quad - \sum_{t=1}^T \alpha_t^\top \big[ -(B_t (u_0 + u_t)) + \xi_t - e_{2t} \big] - \sum_{t=1}^T \beta_t^\top \xi_t.
\end{align}

\paragraph{Step 1: Optimality conditions}  
Set derivatives of the Lagrangian with respect to the primal variables to zero:
\[
\frac{\partial \mathcal{L}}{\partial u_0} = A^\top A u_0 - \sum_{t=1}^T B_t^\top \alpha_t = 0 
\quad \Rightarrow \quad u_0 = (A^\top A)^{-1} \sum_{t=1}^T B_t^\top \alpha_t.
\]
\[
\frac{\partial \mathcal{L}}{\partial u_t} = \mu_1 A_t^\top A_t u_t - B_t^\top \alpha_t = 0 
\quad \Rightarrow \quad u_t = \frac{1}{\mu_1} (A_t^\top A_t)^{-1} B_t^\top \alpha_t.
\]
\[
\frac{\partial \mathcal{L}}{\partial \xi_t} = c_1 e_{2t} - \alpha_t - \beta_t = 0 \quad \Rightarrow \quad 0 \le \alpha_t \le c_1 e_{2t}.
\]

\paragraph{Step 2: Dual problem}  
Substitute the expressions for $u_0$ and $u_t$ back into the Lagrangian to eliminate the primal variables. This yields the following single QPP in $\alpha = [\alpha_1^\top, \dots, \alpha_T^\top]^\top$:
\begin{align}
	\max_{\alpha} \quad 
	& -\frac{1}{2} \alpha^\top \left( Q + \frac{T}{\mu_1} P \right) \alpha + e_2^\top \alpha, \\
	\text{s.t.} \quad 
	& 0 \le \alpha \le c_1 e_2,
\end{align}
where 
\[
Q = B (A^\top A)^{-1} B^\top, \quad 
P = \mathrm{blkdiag}(P_1, \dots, P_T), \quad 
P_t = B_t (A_t^\top A_t)^{-1} B_t^\top, \]
\[
e_2 = [e_{21}^\top, \dots, e_{2T}^\top]^\top.
\]

\paragraph{Step 3: Recovering primal variables}  
After solving the dual problem, the primal variables are obtained as:
\begin{align}
	u_0 &= -(A^\top A)^{-1} B^\top \alpha, \\
	u_t &= -(A_t^\top A_t)^{-1} B_t^\top \alpha_t.
\end{align}

\paragraph{Step 4: Negative class hyperplanes}  
Similarly, for the negative class, the dual problem is:
\begin{align}
	\max_{\alpha^*} \quad 
	& -\frac{1}{2} {\alpha^*}^\top \left( R + \frac{T}{\mu_2} S \right) \alpha^* + e_1^\top \alpha^*, \\
	\text{s.t.} \quad 
	& 0 \le \alpha^* \le c_2 e_1,
\end{align}
with
\[
\alpha^* = \begin{bmatrix} {\alpha_1^*}^\top, \dots, {\alpha_T^*}^\top \end{bmatrix}^\top, \quad 
R = A (B^\top B)^{-1} A^\top, \quad 
S = \mathrm{blkdiag}(S_1, \dots, S_T), \] 
\[
S_t = A_t (B_t^\top B_t)^{-1} A_t^\top,
\]
and the corresponding primal solutions:
\begin{align}
	v_0 &= -(B^\top B)^{-1} A^\top \alpha^*, \\
	v_t &= -\frac{T}{\mu_2} (B_t^\top B_t)^{-1} A_t^\top \alpha_t^*.
\end{align}

This derivation explicitly shows how the dual DMTSVM problems are obtained from the Lagrangian, while preserving the shared and task-specific hyperplane structure across multiple tasks.

\begin{remark}
	Throughout this study, a regularization adjustment is incorporated into the computation of inverse matrices such as $(A^{T}A)$, in order to overcome potential singularity and ill-conditioning issues. Rather than directly evaluating the inverse of $(A^{T}A)$, a small positive constant $\delta$ is added to the diagonal entries, yielding the modified form $(A^{T}A + \delta I)^{-1}$. This modification is introduced to alleviate the aforementioned difficulties while maintaining the structural integrity of the data. By perturbing the matrix with $\delta$, the risk of singularity is avoided and the inversion process remains numerically stable and well-conditioned, even when the original matrix is nearly singular or poorly conditioned.
\end{remark}

As a consequence of solving problems (18) and (19), the hyperplanes corresponding to each task can be expressed as $(u_{0}+u_{t})$ and $(v_{0}+v_{t})$. Subsequently, for a new data point $x$ belonging to the $t$-th task, its class label $k \in \{+1, -1\}$ is assigned according to the following decision rule:
\begin{equation}
	\text{class } k = \arg\min_{k=1,2} \frac{\lvert x^{T}w_{kt} + b_{kt}\rvert}{\lVert w_{kt}\rVert}.
\end{equation}
It is evident that a linear classifier is inadequate when the training data are not linearly separable. To address this challenge, the kernel method is employed. Specifically, we introduce the kernel 
function $K(\cdot,\cdot)$ and define:
\[
D = \big[ A_1^{T}, B_1^{T}, A_2^{T}, B_2^{T}, \ldots, A_T^{T}, B_T^{T} \big]^{T},
\]
\[
\bar{A} = \big[ K(A, D^{T}), e_1 \big], \quad 
\bar{A_t} = \big[ K(A_t, D^{T}), e_{1t} \big],
\]
\[
\bar{B} = \big[ K(B, D^{T}), e_2 \big], \quad 
\bar{B_t }= \big[ K(B_t, D^{T}), e_{2t} \big],
\]

Based on these definitions, the nonlinear MTSVM formulations are expressed as:

\begin{equation}
	\begin{aligned}
		\min_{u_{0}, u_{t}, \xi_t} \quad & 
		\frac{1}{2}\| \bar{A}u_{0} \|^2 
		+ \frac{\mu_1}{2T}\sum_{t=1}^{T} \| \bar{A}_t u_{t} \|^2 
		+ c_1 \sum_{t=1}^{T} e_{2t}^{T}\xi_t \\
		\text{s.t.} \quad & -\bar{B_t}(u_{0}+u_{t})+\xi_t \geq e_{2t}, \\
		& \xi_t \geq 0,
	\end{aligned}
	\tag{20}
\end{equation}
and
\begin{equation}
	\begin{aligned}
		\min_{v_{0}, v_{t}, \eta_t} \quad & 
		\frac{1}{2}\| \bar{B}v_{0} \|^2 
		+ \frac{\mu_2}{2T}\sum_{t=1}^{T} \| \bar{B}_t v_{t} \|^2 
		+ c_2 \sum_{t=1}^{T} e_{1t}^{T}\eta_t \\
		\text{s.t.} \quad & \bar{A}_t(v_{0}+v_{t})+\eta_t \geq e_{1t}, \\
		& \eta_t \geq 0.
	\end{aligned}
	\tag{21}
\end{equation}

Here, $c_1, c_2$ are penalty parameters, while $\xi_t, \eta_t$ denote the slack vectors. 
The number of tasks is denoted by $T$, and $\mu_1, \mu_2 > 0$ are task-preference parameters. 

Using Lagrangian duality and the KKT conditions, the dual problems of (20) and (21) can be written as:

\begin{equation}
	\begin{aligned}
		\max_{\alpha_1} \quad & 
		-\frac{1}{2} 
		\alpha_1^{T}
		\Big(Q + \tfrac{T}{\mu_1}P\Big)
		\alpha_1 
		+  \alpha_1^{T} e_2\\
		\text{s.t.} \quad & 0 \leq \alpha_1 \leq c_1 e_2, 
	\end{aligned}
	\tag{22}
\end{equation}
and
\begin{equation}
	\begin{aligned}
		\max_{\alpha_1^*} \quad & 
		-\frac{1}{2} 
		\alpha_1^{*T} 
		\Big(R + \tfrac{T}{\mu_2}S\Big)
		\alpha_1^* 
		+  \alpha_1^{*T}e_1 , \\
		\text{s.t.} \quad & 0 \leq \alpha_1^* \leq c_2 e_1, 
	\end{aligned}
	\tag{23}
\end{equation}

with
\[
Q = \bar{B}(\bar{A}^T\bar{A})^{-1}\bar{B}^T, 
\quad 
P = \mathrm{blkdiag}(P_1,\ldots,P_T),
\]
\[
R = \bar{A} (\bar{B}^T\bar{B})^{-1}\bar{A}^T , 
\quad 
S = \mathrm{blkdiag}(S_1,\ldots,S_T).
\]

Finally, a new data point $x$ in the $t$-th task is classified into $k \in \{+1,-1\}$ according to:
\begin{equation}
	\text{class } k = \arg\min_{k=1,2} 
	\frac{\big\lvert K(x, D^T) w_{kt} + b_{kt}\big\rvert}{\lVert w_{kt}\rVert}.
\end{equation}

\begin{remark}
	Twin Support Vector Machines  solve two smaller QPPs, 
	each corresponding to one class while constraints are defined by patterns of the other class. 
	Because each QPP involves roughly half the data, TWSVM is approximately four times faster than standard SVM. 
	
	In contrast M-SVM, DMTSVM jointly optimizes the hyperplanes of all $T$ tasks, 
	introducing shared components to enable knowledge transfer. 
	While this improves generalization, especially when tasks are related or per-task data are limited, 
	it increases computational complexity: the naive joint solve has complexity $O((Tm)^3)$ 
	compared to $O(T m^3 / 4)$ for independent TWSVMs. 
	Thus, DMTSVM offers statistical benefits at the cost of reduced scalability for large $T$ or $m$.
\end{remark}


\subsection{Recent Extensions of TWSVM for Multi-Task Learning}
In this section, we provide a comprehensive review of recent advancements in MTL models based on TWSVM. This narrative traces the evolution from foundational models to sophisticated extensions, emphasizing how each innovation addresses key limitations such as computational complexity, noise sensitivity, outlier robustness, class imbalance, and task heterogeneity. 

In 2012, The researchers addressed the challenge of improving generalization performance in multitask classification problems, where single-task learning methods fail to leverage shared information across related tasks. They proposed a novel method called DMTSVM \cite{ref31}, which integrates the regularization principle with twin support vector machines to enable simultaneous learning of related tasks. By sharing information across tasks, DMTSVM generates two non-parallel hyperplanes for classification, and experimental results demonstrated significantly better performance and generalization compared to learning each task independently.

In 2019, the researchers tackled the challenge of enhancing the efficiency and scalability of MTL for pattern recognition, where single-task learning methods fail to utilize shared information across tasks, and methods are computationally inefficient for large data sets due to solving dual QPPs. They proposed a novel Multi-task Least Squares Twin Support Vector Machine (MTLS-TWSVM) \cite{ref32}, which simplifies computations by solving two smaller linear equation systems instead of QPPs, enabling faster processing. The method handles linearly inseparable data using the kernel trick and is scalable for large data sets. 

Following this, in 2021, researchers introduced pinball loss to enhance noise insensitivity over DMTSVM, improving robustness in outlier-prone settings. They named the proposed approach Pinball Loss-Based Multi-Task Twin Support Vector Machine (Pin-DMTSVM) \cite{ref33}. Optimized via dual QPPs with a safe screening rule (SSR) to pre-assign inactive instances, it reduced computational time while maintaining classification accuracy, as demonstrated on six benchmark and seven image data sets. However, it did not address class imbalance or incorporate prior knowledge like Universum data, unlike SVM-based models like U-MTLSVM.

Also, in 2021, the authors tackled the challenge of improving classification performance in MTL for pattern classification, where traditional TSVMs focus on single-task learning  and ignore task relationships, and existing MTL methods apply uniform penalties to misclassified samples, leading to potential overfitting. They proposed a Rough Margin-based Multi-task $\nu$-Twin Support Vector Machine (Rough MT-$\nu$-TSVM) \cite{ref34}, which assigns different penalties to misclassified samples based on their positions, leveraging the strengths of rough set theory and $\nu$-TSVM \cite{ref35}. This approach explores shared information across tasks while preserving individual task characteristics, enhancing generalization and avoiding overfitting. 

The researchers in 2022 resolved the challenge of handling outliers in multitask classification, where the hinge loss in DMTSVM causes hyperplanes to shift toward outliers, reducing generalization performance. They proposed a novel Ramp Loss-based Multi-task Twin Support Vector Machine (RaMTTSVM) \cite{ref36}, which uses ramp loss to limit the impact of outliers and solves non-convex problems via the CCCP method, improving testing accuracy. To tackle the computational complexity, they introduced a Multi-parameter Safe Acceleration (MSA) rule, leveraging the sparsity of ramp loss to eliminate inactive samples before solving, thus reducing problem size while preserving accuracy. 

Also, in 2022, The researchers handled the challenge of enhancing MTL for classification tasks, particularly in the presence of noisy data, where traditional single-task learning  methods may underperform due to their inability to leverage shared information across tasks. They proposed a Robust Multi-task Least Squares Twin Support Vector Machine (RMTLS-TSVM) \cite{ref37}, inspired by the MTLS-TWSVM \cite{ref32}. This model introduces an error factor to effectively handle noise, making it robust against data imperfections. By solving two linear systems instead of complex quadratic programming problems, RMTLS-TSVM is computationally efficient and easy to implement, enabling its application to large-scale and real-world data sets. 

Furthermore, in 2022, The researchers tackled the challenge of applying the computationally intensive DMTSVM to large-scale problems, where its high computational cost limits scalability despite leveraging shared information across tasks for better performance than single-task TSVM. They proposed two safe sample screening rules (SSRC-DMTSVM and SSRR-DMTSVM) \cite{ref38} to identify and eliminate inactive instances before solving the optimization problem, significantly reducing computational cost. These rules are "safe" as they preserve the exact same solution as the original problem, ensuring no loss in accuracy. Additionally, they combined these rules with a Dual Coordinate Descent Method (DCDM) to further accelerate computation. 

Also, the researchers in 2022, addressed the challenge of classifying imbalanced data sets, where standard SVMs often perform poorly on the minority class by mistaking it for noise and struggle with data sets containing noise and outliers. They proposed Intuitionistic Fuzzy Twin Support Vector Machines for Imbalanced Data (IFTSVM-ID) \cite{ref39}, a method that enhances robustness against imbalanced classes, noise, and outliers. IFTSVM-ID employs a reasonable weighting strategy to balance the influence of imbalanced classes and a margin-based technique to mitigate the impact of noise and outliers. It formulates both linear and non-linear kernel functions to construct two non-parallel hyperplanes for classification. 

In 2023, the authors tackled the challenge of classifying imbalanced data sets within a MTL framework, where existing MTL methods often perform poorly on imbalanced data common in real-world scenarios. They proposed a novel method called Multi-task Twin Spheres Support Vector Machine with Maximum Margin (MTMMTSVM) \cite{ref40}, which constructs two homocentric hyper-spheres for each task and employs the maximum margin principle to separate majority and minority samples. By exploring shared commonalities and individual task characteristics, MTMMTSVM solves a linear programming problem (LPP) and a smaller QPP.
In 2024, The authors addressed the challenge of enhancing the performance of MTL for classification by leveraging additional information beyond target task data, particularly when traditional MTL models fail to utilize non-target data from the same domain. They proposed a Multi-task Twin Support Vector Machine with Universum Data (UMTSVM) \cite{ref41}, incorporating Universum data that does not belong to any class but shares the same domain as the target data—as prior knowledge to improve generalization. Two approaches were developed: the first, UMTSVM, formulates a dual quadratic programming problem (QPP) to solve the classification task, while the second, Least Squares UMTSVM (LS-UMTSVM), simplifies the solution to two systems of linear equations, significantly reducing computational complexity and enhancing speed.

Moreover, In 2024, the authors addressed the challenge of accurately classifying complex human activities from sensor data in human activity recognition (HAR), which is critical for applications like healthcare and sports analysis but complicated by high-dimensional and noisy data. They proposed a novel method combining Particle Swarm Optimization (PSO) for feature optimization with a Non-linear Multi-task Least Squares Twin Support Vector Machine (NMtLSSVM) \cite{ref42}.  NMtLSSVM, integrated into a multitask learning framework, simultaneously learned multiple related tasks, sharing information to enhance generalization and robustness compared to single-task models.

Furthermore, in 2024, the authors addressed the challenge of accurately recognizing human activities from sensory data for applications such as remote monitoring, where computational efficiency and model performance are critical. They proposed a Grid Search Based Nonlinear Multi-task Least Squares Twin Support Vector Machine (GMLT-SVM) \cite{ref43}, which integrates grid search for hyperparameter tuning with a nonlinear kernel within a multitask learning framework. This approach optimizes hyperparameters to enhance model efficacy while reducing computation time compared to traditional methods. By leveraging shared information across multiple tasks, GMLT-SVM improves generalization and robustness. 

In 2025, The researchers tackled the challenge of improving the robustness of matrix classifiers against noisy signals in pattern classification, particularly for applications like mechanical fault diagnosis where traditional classifiers are sensitive to noise. They proposed a Multitask Twin Pinball Kernel Matrix Classifier (MTTPKMC) \cite{ref44}, which integrates a Feature Mapping Kernel (FMK) to preserve the structural information of matrix samples and a pinball constraint term to reduce noise sensitivity. By incorporating a multitask learning (MTL) framework, MTTPKMC leverages correlation information across tasks to enhance classification performance. The method constructs two non-parallel hyperplanes for each task, improving robustness and generalization over single-task approaches. 

Also, In 2025, the researchers tackled the challenge of improving the generalization performance of multitask learning (MTL) for classification, where existing methods like the DMTSVM are computationally intensive due to solving quadratic programming problems. They proposed an IMTLS-TSVM \cite{ref45}, building on the IMTLS-TSVM \cite{ref32}. Unlike DMTSVM, IMTLS-TSVM solves two linear systems of equations, significantly reducing computational time. The model enhances generalization by incorporating regularization terms in its objective function, adhering to the empirical risk minimization principle to control model complexity and prevent overfitting.

These extensions reflect a shift toward robust, knowledge-infused TWSVM models, addressing scalability, noise, and real-world applicability in MTL.

Table \ref{tab:mtl-twsvm} presents representative models in multi-task learning based on TWSVM, outlining their key features and optimization strategies. The listed models demonstrate significant methodological advancements, such as error handling through robust loss functions, safe screening rules for computational efficiency, incorporation of Universum and fuzzy data, and hybrid approaches integrating evolutionary algorithms for feature optimization. Additionally, kernel tricks, decomposition methods, and non-convex optimization have been employed to enhance scalability and robustness against noisy or imbalanced data.

\begin{longtable}{l l p{5cm} p{6cm}} 
	\caption{Representative Models in Multi-Task Learning based on TWSVM }\label{tab:mtl-twsvm} \\
	\hline
	Model & Year & Key Features & Optimization Approach \\
	\hline
	\endfirsthead
	\hline
	Model & Year & Key Features & Optimization Approach \\
	\hline
	\endhead
	\hline \multicolumn{4}{r}{{Continued on next page}} \\
	\endfoot
	\hline
	\endlastfoot
	
	\cite{ref31}& 2012 & Extends TSVM to multi-task learning, generates two nonparallel hyperplanes per task, uses shared representation to leverage task relatedness, solves smaller QPPs for faster computation compared to standard SVM & Solves a pair of smaller QPPs per task with regularization, using linear kernel, optimized for efficiency \\
	\cite{ref32} & 2019 & Extends LSTWSVM to multi-task learning, solves linear equations instead of QPPs, uses two nonparallel hyperplanes per task, leverages shared task information, handles linear inseparable data with kernel trick, suitable for large-scale datasets & Solves a pair of small linear equations per task with Gaussian kernel, tuned by parameters $\lambda$, $\gamma$, $\rho$ via grid search, optimized for parallel/distributed computing \\
	\cite{ref33}& 2021 & Enhances DMTSVM with pinball loss for noise insensitivity, generates two nonparallel hyperplanes per task, uses safe screening rule to reduce computational cost & Solves two smaller QPPs per task using dual formulation, employs Gaussian kernel, uses SSR to screen inactive samples \\
	\cite{ref34} & 2021 & Extends rough $\nu$-TSVM to multi-task learning, assigns different penalties to misclassified samples based on position (lower margin, upper margin, boundary), leverages task correlations to improve generalization, uses SDM for large-scale acceleration & Solves two QPPs per task with rough margin constraints, employs Gaussian kernel, tunes parameters via grid search, uses SMO-type decomposition method (SDM) for fast solving of subproblems with two variables \\
	\cite{ref36}& 2022 & Uses ramp loss to limit maximum penalty for outliers, enhances noise robustness, generates two nonparallel hyperplanes per task, employs multi-parameter safe acceleration (MSA) to reduce computational cost, maintains identical solutions & Solves non-convex problem using CCCP with series of convex QPPs, employs Gaussian kernel, tunes parameters via grid search, uses MSA to screen inactive samples for multiple parameters and CCCP iterations \\
	\cite{ref37} & 2022 & Proposes robust multi-task least squares twin SVM (RMT-LSTSVM) for classification, introduces error factor to handle noise, extends multi-task LSTSVM to improve robustness, supports nonlinear data via kernel trick, efficient and scalable for large real-world datasets & Solves quadratic programming problems for twin hyperplanes, employs kernel functions for nonlinear patterns \\
	\cite{ref38} & 2022 & Enhances DMTSVM with safe screening rules (SSRC for parameter $c$, SSRR for parameter $\rho$), identifies and deletes inactive samples before solving, maintains identical solutions, combines with DCDM for further acceleration & Solves two smaller QPPs per task using dual formulation, employs Gaussian kernel, uses SSRC/SSRR for screening, applies DCDM for faster solving \\
	\cite{ref39} & 2022 & Handles imbalanced datasets with noises/outliers using intuitionistic fuzzy sets, assigns weights based on imbalance ratio, maximizes margin to reduce noise impact, supports linear and non-linear kernels & Solves two stable positive definite QPPs with regularization term, employs linear or Gaussian kernel, tunes parameters via 5-fold cross-validation and grid search \\
	\cite{ref40} & 2023 & Extends MMTSVM to multi-task learning for imbalanced data, constructs two homocentric hyper-spheres per task, maximizes margin to focus on minority samples, uses LPP and smaller QPP for efficiency, integrates fast decomposition method (DM) for large-scale problems & Solves one LPP and one smaller QPP per task with Gaussian kernel, tunes parameters $\mu$, $v_1$, $v_2 \in [0.1,0.9]$, employs DM for acceleration, uses 5-fold cross-validation and grid search \\
	\cite{ref41}& 2024 & Extends TSVM to multi-task learning with Universum data as prior knowledge, generates two nonparallel hyperplanes per task, incorporates non-target task data for better generalization, offers dual QPP (UMTSVM) and least-squares (LS-UMTSVM) solutions for faster computation & UMTSVM: Solves quadratic programming problem via dual formulation; LS-UMTSVM: Solves two systems of linear equations per task with Gaussian kernel \\
	\cite{ref42} & 2024 & Combines PSO for feature optimization with non-linear multi-task least squares twin SVM for HAR, reduces dimensionality by 60\% while maintaining accuracy, handles multi-class activities from sensor data, achieves 97.8\% (UCI-HAR) and 98.5\% (WISDM) accuracy & PSO for feature selection (population size 50-1000, iterations up to 1000), NMtLSSVM solves linear equations per task with Gaussian kernel \\
	\cite{ref44} & 2025 & Extends twin SVM to multi-task matrix classification with pinball loss for noise insensitivity, uses feature mapping kernel (FMK) to preserve structural information in high-dimensional space, leverages task correlations via MTL framework for improved generalization, suitable for mechanical fault diagnosis & Solves two smaller QPPs per task with pinball constraint and FMK, employs Gaussian kernel \\
	\cite{ref45} & 2025 & Enhances MTLS-TSVM with empirical risk minimization, solves two linear equations per task, includes regularization terms to prevent overfitting, generates two nonparallel hyperplanes, leverages shared task information for better generalization & Solves two systems of linear equations per task with regularization terms, uses linear or Gaussian kernel, tuned by grid search \\
\end{longtable}

Table \ref{tab:mtl-twsvm-eval} summarizes the evaluation factors commonly used to benchmark MTL-TWSVM models, including accuracy, running time, MAP, speedup, robustness to noise, precision-recall-F-score metrics. 
\begin{table}
	\centering
	\caption{Evaluation Factors in Multi-Task Learning based on TWSVM}
	\label{tab:mtl-twsvm-eval}
	\begin{tabular}{|c|p{2cm}|p{2cm}|p{2cm}|p{2.5cm}|p{2.5cm}|}
		\hline
		\textbf{Model} & \textbf{Accuracy} & \textbf{Running Time} & \textbf{Speedup} & \textbf{Robustness to Noise} & \textbf{Precision, Recall, F-Score} \\
		\hline
		\cite{ref31} & $\checkmark$ &  &  &  &  \\
		\cite{ref32}& $\checkmark$ & $\checkmark$ &  &  & $\checkmark$ \\
		\cite{ref33}& $\checkmark$ & $\checkmark$ & $\checkmark$ & $\checkmark$ &  \\
		\cite{ref34} & $\checkmark$ & $\checkmark$ &  &  &  \\
		\cite{ref36} & $\checkmark$ &  &  & $\checkmark$ &  \\
		\cite{ref37} & $\checkmark$ &  &  & $\checkmark$ &  \\
		\cite{ref38} & $\checkmark$ &  & $\checkmark$ & $\checkmark$ &  \\
		\cite{ref39}& $\checkmark$ &  &  & $\checkmark$ &  \\
		\cite{ref40}& $\checkmark$ & $\checkmark$ & $\checkmark$ &  &  \\
		\cite{ref41} & $\checkmark$ & $\checkmark$ &  &  &  \\
		\cite{ref42}  & $\checkmark$ &  &  &  &  \\
		\cite{ref44} & $\checkmark$ &  &  & $\checkmark$ & $\checkmark$ \\
		\cite{ref45} & $\checkmark$ & $\checkmark$ &  &  & $\checkmark$ \\
		\hline
	\end{tabular}
\end{table}
\section{Comparison between Multi-Task SVM and Multi-Task TWSVM}

In multi-task learning, SVM and TWSVM have been extended to leverage inter-task relationships through parameter sharing. M-SVM employs a direct parameter-sharing scheme, where the weight vector of each task is decomposed into a global shared component and a task-specific deviation. This structure encourages knowledge transfer across tasks by aligning task hyperplanes to a common direction while retaining task-specific flexibility.

In contrast, the DMTSVM extends the twin-hyperplane framework of TWSVM to the multi-task context. Each task is associated with two hyperplanes per class, each consisting of shared and task-specific components. This design maintains the class-oriented separation characteristic of TWSVM while enabling inter-task knowledge transfer. Although DMTSVM generally exhibits improved generalization, particularly in imbalanced or heterogeneous task settings, it introduces additional parameters and may incur higher computational costs when optimized jointly.

\begin{table}
	\centering
	\caption{Comparison between M-SVM and DMTSVM}
	\begin{tabular}{p{2cm} p{5cm} p{5cm}}
		\toprule
		\textbf{Aspect} & \textbf{ M-SVM} & \textbf{DMTSVM} \\
		\midrule
		\textbf{Formulation} & 
		Each task weight vector is expressed as 
		$\mathbf{w}_t = \mathbf{w}_0 + \mathbf{v}_t$, where $\mathbf{w}_0$ is the shared component and $\mathbf{v}_t$ is task-specific. A joint convex QPP is solved with regularization parameters $\mu$ and $C$. & 
		Extends TWSVM by introducing shared and task-specific components for each class hyperplane, e.g., $\mathbf{u}_0 + \mathbf{u}_t$ and $\mathbf{v}_0 + \mathbf{v}_t$. Joint optimization is controlled by parameters $\mu_1, \mu_2, c_1, c_2$. \\
		\midrule
		\textbf{Geometric property} & 
		A single global direction is enforced across tasks, with small deviations per task. Margin-based formulation encourages alignment to a shared hyperplane. & 
		Retains TWSVM’s two nonparallel hyperplanes per class, now decomposed into shared and task-specific parts, allowing class-wise separation with inter-task knowledge sharing. \\
		\midrule
		\textbf{Computational complexity} & 
		Solving the joint QPP can be expensive: $O((T m)^3)$, limiting scalability when $T$ or $m$ is large. & 
		Independent TWSVM training per task has cost $\sim O(T m^{3}/4)$, but DMTSVM joint optimization may reach $O((T m)^3)$. \\
		\midrule
		\textbf{Scalability} & 
		Limited scalability in joint settings; requires decomposition, approximation, or distributed optimization for large $T$ or $m$. & 
		Scalability is challenged due to the twin-hyperplane structure, though linear or least-squares variants significantly improve efficiency. \\
		\midrule
		\textbf{Generalization} & 
		Strong margin-based generalization, particularly effective when tasks are related; parameter $\mu$ balances knowledge transfer versus task independence. & 
		Superior performance in imbalanced or heterogeneous class distributions; twin-hyperplane modeling enhances class-specific generalization. \\
		\midrule
		\textbf{Advantages} & 
		Solid theoretical foundation in margin theory and RKHS; straightforward kernelization; effective for homogeneous tasks. & 
		Better handling of class imbalance and heterogeneous tasks; potential for faster training in LS/approximate variants; preserves TWSVM’s class-wise flexibility. \\
		\midrule
		\textbf{Limitations} & 
		High computational burden for large-scale joint problems; risk of negative transfer if tasks are poorly aligned. & 
		Requires tuning multiple parameters; joint optimization may be computationally demanding; base formulations lack direct multi-class extensions unless further generalized. \\
		\midrule
		\textbf{Optimization methods} & 
		Typically solved via SMO or standard QP solvers; kernel methods widely supported. & 
		Variants employ safe screening, least-squares reformulations, or alternative losses (e.g., pinball, ramp) to improve stability and reduce computational cost. \\
		\bottomrule
	\end{tabular}
	\label{tab:multi-task-comparison}
\end{table}

Both approaches offer principled frameworks for exploiting task relatedness. M-SVM provides a strong margin-based theoretical foundation and is well-suited for moderately sized, homogeneous task sets. DMTSVM and its variants are particularly advantageous for heterogeneous or imbalanced tasks, though at the cost of increased parameterization and higher computational overhead in joint optimization.


\section{Conclusion and Future Directions}

This chapter has provided a comprehensive survey of multi-task learning (MTL) methodologies grounded in the principles of Support Vector Machines (SVMs) and Twin Support Vector Machines (TWSVMs). We began by establishing the foundational paradigms of MTL and detailing the core mechanics of standard SVMs and TWSVMs, highlighting the latter's computational efficiency gained through dual, non-parallel hyperplanes. Our in-depth review traced the evolution of both SVM-based and TWSVM-based MTL models, demonstrating how they leverage shared representations, parameter-based coupling, and robust loss functions to enhance generalization across related tasks.

The comparative analysis reveals a clear trade-off. SVM-based MTL approaches, such as the foundational M-SVM, offer a strong theoretical foundation rooted in statistical learning theory and are particularly effective for homogeneous task sets. In contrast, TWSVM-based methods, like DMTSVM and its extensions, inherit the inherent efficiency of TWSVMs and exhibit superior performance in handling imbalanced data and heterogeneous task distributions due to their class-specific modeling. While joint optimization in multi-task frameworks can incur significant computational cost, the field has responded with innovative strategies, including least-squares formulations, safe sample screening rules, and functional iterative methods, to improve scalability and practicality.

Despite these significant advancements, the field of margin-based MTL remains ripe with opportunities for further exploration. Based on our survey, we identify several key directions for future research:
\begin{itemize}
\item \textbf{1. Scalability to Large-Scale and Many-Task Settings}

While least-squares and screening rules have alleviated some computational burdens, scaling the joint optimization of M-SVM and DMTSVM to hundreds or thousands of tasks (the ``many-task'' problem) remains a challenge. Future work should focus on:
\begin{itemize}
    \item Developing efficient distributed and parallel optimization algorithms tailored for these models.
    \item Exploring federated learning frameworks for margin-based MTL, enabling model training across decentralized data sources without sharing raw data.
    \item Creating more advanced, dynamic safe screening rules that can identify and eliminate non-support vectors across all tasks simultaneously.
\end{itemize}

\item \textbf{2. Integration with Deep Learning and Hybrid Architectures}

The interpretability of kernel-based models and the flexibility of deep learning are not mutually exclusive. Promising avenues include:
\begin{itemize}
    \item Designing hybrid architectures where deep neural networks serve as adaptive feature extractors, and the final layers are composed of SVMs or TWSVMs for robust, margin-based classification.
    \item Using deep networks to learn the optimal kernel function or the task-coupling matrix for SVM-based MTL, moving beyond pre-defined kernels like RBF or polynomial.
\end{itemize}

\item \textbf{3. Advancements in Multi-Class and Structured Output Settings}

Most MTL extensions of TWSVM are inherently binary classifiers. Real-world problems often require multi-class classification or more complex structured outputs (e.g., sequences, graphs).
\begin{itemize}
    \item Future work should generalize the twin-hyperplane concept to native multi-class MTL formulations, building on frameworks like ``one-versus-one-versus-rest.''
    \item Exploring MTL extensions of SVMs and TWSVMs for structured output prediction, such as in natural language processing or bioinformatics, would significantly broaden their applicability.
\end{itemize}

\item \textbf{4. Explainability and Formal Interpretability}

The ``interpretability'' of SVMs is often cited but rarely formalized in the MTL context. Future studies should:
\begin{itemize}
    \item Develop methods to visualize and quantify the contribution of the shared component $\mathbf{w}_0$ versus task-specific components $\mathbf{v}_t$.
    \item Create model-agnostic or intrinsic techniques to explain which features are leveraged for knowledge transfer and which are task-specific, enhancing trust and debuggability in critical applications like healthcare and finance.
\end{itemize}
\end{itemize}
In conclusion, SVM and TWSVM provide a powerful, theoretically sound, and interpretable foundation for multi-task learning. By addressing the research challenges outlined above, the next generation of margin-based MTL models can achieve new levels of scalability, robustness, and intelligence, solidifying their role as indispensable tools in the machine learning landscape.

%

\eject

\end{document}